\documentclass{article}

\PassOptionsToPackage{table}{xcolor}

\usepackage[a4paper,margin=1in]{geometry}
\usepackage{mathptmx}            
\usepackage[T1]{fontenc}
\usepackage[utf8]{inputenc}
\usepackage{microtype}
\usepackage{inconsolata}
\usepackage{graphicx}
\usepackage{booktabs}
\usepackage{colortbl}
\usepackage{amsmath}
\usepackage{xspace}
\usepackage{tcolorbox}
\tcbuselibrary{breakable}
\usepackage{fvextra}
\usepackage{pgfplots}
\pgfplotsset{compat=1.16}
\usepackage{tikz}
\usepackage{tabularx}
\usepackage{multirow}
\usepackage{siunitx}
\usepackage{caption}
\usepackage[round]{natbib}
\usepackage[colorlinks=true,linkcolor=blue!45!black,citecolor=blue!45!black,urlcolor=blue!45!black]{hyperref}

\DefineVerbatimEnvironment{PromptVerbatim}{Verbatim}{
    breaklines=true,
    breakanywhere=true,
    breaksymbolleft={},
    breaksymbolright={},
    fontsize=\small,
    baselinestretch=0.95,
    breakindent=0pt
}

\newcommand{\agentname}{Explyt AI Agent\xspace}
\newcommand{\abbrevagentname}{EAA\xspace}
\newcommand{\githuburl}{\url{https://github.com/agent-lens/agent-lens-bench}}

\usepackage{authblk}
\date{\today}

\begin{document}

\title{AgentLens: Production-Assessed Trajectory Reviews for Coding Agent Evaluation}

\author[1]{Andrey~Podivilov$^{\ast}$}
\author[1]{Vadim~Lomshakov$^{\ast,\dagger}$}
\author[1]{Sergey Savin$^{\ast}$}
\author[1]{Matvei~Startsev}
\author[1]{Roman~Pozharskiy}
\author[1]{Maksim~Parshin}
\author[2,3]{Sergey~Nikolenko}
\affil[1]{Explyt}
\affil[2]{St. Petersburg Department of the Steklov Institute of Mathematics}
\affil[3]{St. Petersburg State University}

\maketitle

\begin{abstract}
We present AgentLens, a production-assessed benchmark for interactive code agents.
Most code-agent benchmarks reduce a run to a single bit---did the task pass?---but the people who actually use these agents experience the entire trajectory: how the agent follows instructions, uses its tools, verifies its own work, recovers from mistakes, and talks to them along the way.
AgentLens evaluates that whole trajectory.
It pairs formal verification, where an objective check exists, with LLM-written trajectory reviews and side-by-side comparisons, so that each run yields a readable explanation of {why} the score is what it is.
This makes AgentLens useful for more than ranking models: we use it to diagnose model behavior, compare successive versions of our own agent, and catch product regressions in a nightly evaluation pipeline.
We release the benchmark as open source at \githuburl.
\end{abstract}

\thanks{\small$^{\ast}$Equal contribution. $^{\dagger}$Corresponding author: \texttt{vadim.lomshakov@gmail.com}}

\section{Introduction}

Most coding agent benchmarks ask one question: did the agent finish the task~\citep{jimenez2024swebench, terminalbench}?
For a research leaderboard, a single pass/fail bit is often enough.
But for a coding assistant that developers use all day, most of what matters is not captured by that bit.
A run can pass a narrow check and still be unreliable, overconfident, unpleasant to work with, or subtly misaligned with what the user actually asked for.
Conversely, a run that fails the final check may have done plenty of useful things on the way there: localized the bug correctly, used its tools safely, recovered gracefully from an error and so on.

The problem is most important for tasks that have no natural binary outcome, such as, e.g., writing project documentation. A single success flag cannot say much about it: the useful question is not whether a file appeared on disk, but whether the document is accurate, appropriately scoped, readable, grounded in the actual codebase, and helpful to the readers.

Our proposed benchmark, \emph{AgentLens}, is built to keep this structure visible, evaluating complete trajectories rather than final states alone. A \emph{trajectory} is the full record of a session: messages from the (simulated) user, the agent's replies, its tool calls, file edits, command executions, verification attempts, the final response, and the final state of the repository. AgentLens treats this record as the primary object of evaluation and uses it to measure quality closer to the way a user would perceive it. The benchmark uses textual judge reviews as the main artifacts, while formal verification supplies complementary objective checks wherever such checks exist.

We built AgentLens because the benchmarks we had did not match what we needed to know: some lacked realistic user interaction, some collapsed everything to a binary or final-state outcome, and some used tasks that were either too simple or too far from the day-to-day work of a coding assistant. We wanted something that evaluates production-like interactive work and then {explains} how agents differ.

Our main contributions in this work are:
\begin{itemize}
    \item a production-assessed benchmark for interactive code agents;
    \item a trajectory-level evaluation protocol that combines formal verification with LLM judge reviews;
    \item side-by-side reviews for model comparison, feature evaluation, and regression analysis;
    \item an open-source release of the benchmark code at \githuburl;
    \item a nightly evaluation pipeline that detects regressions in an actively developed coding agent.
\end{itemize}

The rest of the paper is organized as follows. Section~\ref{sec:related} places AgentLens in the context of prior code-agent, tool-use, and LLM-as-judge benchmarks. Section~\ref{sec:agentlens} is the core of the paper: it describes the task set and simulated users, the trajectory review protocol and its metrics, the quality index, side-by-side reviews, the reliability and bias checks, and the production pipeline that runs all of this nightly. Section~\ref{sec:evaluation} puts the benchmark to work, with a leaderboard of current models (Section~\ref{sec:leaderboard}), a qualitative analysis of how they differ and what the reviews catch (Section~\ref{sec:qualitative}), and a correlation study against public benchmarks (Section~\ref{sec:aa-correlation}). Sections~\ref{sec:limitations} and~\ref{sec:conclusion} conclude the paper with limitations and future work.

\section{Related Work}
\label{sec:related}

AgentLens builds on three lines of work: code-agent benchmarks, interactive tool-use environments, and LLM-as-judge evaluation. The main novelty of AgentLens is in the object it evaluates, which is a complete, production-like trajectory including the user request, the agent's messages, its tool calls, file edits, verification attempts, recovery behavior, final answer, and final repository state.

\paragraph{Code and software-engineering benchmarks.}
Early code generation benchmarks evaluate isolated programming tasks, usually with unit tests as the success criterion.
Repository-level benchmarks move closer to real software engineering. SWE-bench evaluates whether language models can resolve real GitHub issues by modifying repositories and passing tests~\citep{jimenez2024swebench}, and SWE-agent shows that the agent--computer interface is itself an important factor in automated software engineering~\citep{yang2024sweagent}.
These benchmarks are close to AgentLens in their focus on coding agents and realistic repositories, but they primarily measure final-state task resolution.
AgentLens keeps formal verification where it is available and complements it with trajectory-level reviews that capture instruction following, tool discipline, validation behavior, recovery from errors, and user-facing communication.

\paragraph{Interactive and tool-using agents.}
A parallel line of work evaluates LLMs as agents acting over multiple steps in external environments.
AgentBench studies agent performance across diverse environments~\citep{liu2023agentbench}; ToolLLM evaluates the ability to use large collections of real-world APIs~\citep{qin2023toolllm}; WebArena and OSWorld introduce realistic web and computer-control environments~\citep{zhou2024webarena,xie2024osworld}.
Most relevant to our interaction model, $\tau$-bench evaluates tool-agent-user interaction in real-world domains using a simulated user with a role and a hidden goal~\citep{yao2024taubench}.
AgentLens borrows this idea for coding assistant evaluation: user behavior is generated by another LLM conditioned on a persona and a task objective.
This lets us evaluate production-like interactive coding workflows not only by whether the agent completes the repository task, but also by how it behaves with the user throughout the session. Each recorded trajectory can be reused for flexible comparisons across agents and quality dimensions.

\paragraph{Terminal and production-agent benchmarks.}
Terminal-Bench evaluates AI agents in terminal environments on tasks spanning software engineering, system administration, security, and data science~\citep{terminalbench}.
AgentLens also benchmarks CLI agents on realistic tasks drawn from everyday developer workflows, where agents must inspect projects, edit files, run tools, and recover from failures, but AgentLens is built mostly for {product assessment}; the goal is, again, not only to measure task resolution but to diagnose user-visible quality in a deployed coding assistant.

\paragraph{LLM-as-judge evaluation.}
Finally, AgentLens relates to LLM-as-judge and pairwise preference evaluation.
MT-Bench and Chatbot Arena show how LLM judges and side-by-side comparisons can support scalable evaluation of open-ended assistant behavior~\citep{zheng2023judging}.
AgentLens uses LLM judges to score trajectories along metric-specific dimensions and, moreover, supplements the scores with written reviews that lay out the evidence behind them.
These reviews make it possible to tell whether a low score came from poor tool use, incomplete verification, misleading final claims, weak instruction following, or simply an unpleasant interaction style. AgentLens then summarizes reviews across all trajectories in a run, surfacing the most frequent and important failure modes so developers can prioritize fixes.


\section{AgentLens}
\label{sec:agentlens}

\subsection{Task Set and Runs}
\label{sec:task-set}

AgentLens is designed to be easy to extend.
Adding a task requires two things: a task description for the user simulator and a verification setup for the formal metrics.
In common cases, verification can reuse existing checks---running tests, inspecting repository state, or searching for an expected code change---and for specialized cases one can drop in a task-specific verifier (the available verifiers are listed in Table~\ref{tab:verifier-categories}).
New judge metrics can be added when a benchmark fold targets a specific behavior or when developers want to inspect an additional quality dimension; a test-generation fold, for instance, can score mocking strategy with a test-specific metric instead of general coding-assistant behavior.

We release an initial fold of 16 coding agent scenarios.
Each scenario is run with two user personas, a neutral default user and a toxic user (personas are defined below), which yields 32 trajectories per evaluated agent.
The folds are deliberately compact so that each one can run often in production.
The design follows a simple observation that any developer who has lived with a coding assistant for a week will recognize: after four or five real sessions you can usually tell which agent is nicer to work with, and what its recurring strengths and failure modes are.
AgentLens approximates that judgment by having LLM judges review and score complete interaction trajectories rather than isolated final outputs.
The low run-to-run variance reported in Section~\ref{sec:leaderboard} suggests that this compact setup is, in fact, stable.

\paragraph{Where the tasks come from.}
The released fold is organized around \emph{workflows}: each scenario specifies a task objective together with a sequence of instructions and interaction steps.
We obtained these workflows from interviews in which programmers walked us through work they had done earlier that day or the day before.
We asked follow-up questions to recover the specific actions involved---inspecting code, editing files, searching documentation, running tests, debugging failures, reviewing changes---and we asked when each activity had last occurred, giving more weight to workflows developers had performed recently and treating rarely performed activities as less representative.

In addition to these interview-derived workflows, the fold includes production-derived scenarios constructed from anonymized usage summaries, which complement the interviews with workflows observed in real product use. These scenarios come from chats of users who consented to anonymous data collection. We first converted each conversation into an anonymized task summary and processed only the summaries from then on, extracting coarse tags such as programming language, technology, task type, and application domain. This privacy-preserving usage-analysis step is similar in spirit to \emph{Clio}, which summarizes and clusters real assistant conversations to surface aggregate usage patterns without exposing raw conversations~\citep{tamkin2024clio}. We clustered the tag vectors with $k$-means and inspected large clusters that were not yet represented in the benchmark (cluster examples appear in Appendix~\ref{app:task-clusters}). For each such cluster, a programmer picked a matching open-source project, generated a candidate scenario with the coding assistant, and then revised it for realism, anonymity, and verifiability.

\paragraph{User simulator personas.}
\label{sec:user-personas}
Because the user is itself an LLM, we can vary its temperament and watch how robust an agent is to different interaction styles.
We instantiate the simulator with three personas:
\begin{itemize}
\item the \emph{default} persona is a relaxed user who does not proactively help the agent and intervenes only when there is clear evidence of failure;
\item the \emph{helpful} persona is cooperative: when the agent makes a visible mistake, it briefly points out the issue, suggests a possible correction, or offers concise guidance in a neutral tone;
\item the \emph{toxic} persona is mildly frustrated and occasionally impolite---it may challenge or tease the agent---but it still preserves the underlying task objective and never gives intentionally misleading instructions.
\end{itemize}
Together these let us test whether an agent can finish a task not only with a cooperative user, but also under sparse feedback and mildly adversarial conditions.
The released fold pairs each scenario with the default and toxic personas; the helpful persona is available for studies of agents that benefit from in-loop guidance.

\subsection{Trajectory Review Protocol}
\label{sec:review-protocol}

The unit of evaluation is a full trajectory, evaluated by LLMs. This raises several concerns.

\paragraph{Can we trust an LLM to grade other LLMs?}
This is the first question to ask, and it is a fair one. In our experience, human annotation is not the obvious gold standard it has been often assumed to be for long coding trajectories: each item requires reading a lengthy interaction trace and applying a multi-page rubric, which is exhausting and error-prone. Early in the project we ran some informal internal pairwise-annotation experiments, and our consistent impression there was that agreement between an LLM judge and our human annotators was at least as high as agreement among the annotators themselves; this is especially plausible given how taxing these trajectories are to grade by hand. We note that those experiments were preliminary and small in size, so we do not claim a conclusive result here and mark a formal agreement study for future work. However, it is directionally consistent with prior LLM-as-judge work, which finds that strong judges can reach human-level agreement with human preferences, comparable to the agreement between two human annotators~\citep{zheng2023judging}.

\paragraph{Formal verification.}
Each trajectory also receives formal checks wherever they apply: tests, regular-expression checks, repository-state checks, and task-specific validators.
These checks are valuable because they give objective evidence about specific requirements.
But they are only one part of the picture: on their own they usually cannot see interaction quality, instruction compliance, misleading validation, unsafe tool use, or partial coverage of a requirement.
Reviewing the full trajectory also blunts reward hacking against a narrow success predicate.
An agent can satisfy a binary check while using brittle shortcuts, skipping required validation, or misreporting the final state; all of that stays visible in the trajectory review even when the check is green.
Table~\ref{tab:verifier-categories} lists the verifier types currently available.

\begin{table}[!t]
\centering
\caption{Formal verifier categories and their descriptions. A scenario passes formal verification only if every one of its verifiers passes; the run-level report also records the fraction of individual verifiers that passed.}
\label{tab:verifier-categories}
\small
\begin{tabular}{@{}p{0.12\textwidth}p{0.28\textwidth}p{0.53\textwidth}@{}}
\toprule
\textbf{Category} & \textbf{Verifier} & \textbf{Description} \\
\midrule
Repository state checks & {\small\texttt{NoChangesVerifier}} & Ensures no files outside an explicitly allowed set were modified, guarding against unintended side effects. \\
& {\small\texttt{YesChangesVerifier}} & Asserts that a specified set of target files was modified, confirming the agent acted on the correct locations. \\
& {\small\texttt{ExactFileMatchVerifier}} & Performs line-by-line comparison of project files against reference files; optionally ignores empty lines and verifies no other files were changed. \\
\addlinespace
Regular expression checks & {\small\texttt{ChatRegExpCountVerifier}} & Counts regex matches across all assistant messages in the conversation history. \\
& {\small\texttt{ChatOrToolRegExpCountVerifier}} & Extends {\small\texttt{ChatRegExpCountVerifier}} to also search tool call arguments and tool responses. \\
& {\small\texttt{JavaFileRegExpCountVerifier}} & Counts regex matches within Java source files at a given path, with an option to strip comments before matching. \\
& {\small\texttt{NewFileRegExpVerifier}} & Checks that at least one newly created file has a path matching the given pattern. \\
\addlinespace
Test execution & {\small\texttt{JavaRunTestsVerifier}} & Runs the project test suite via Maven or Gradle, parses output for pass/fail/skip counts, and fails if any test does not pass. \\
& {\small\texttt{JavaTestCoverageVerifier}} & Measures and asserts the minimum test coverage for the specified class in the project. \\
\addlinespace
Command output checks & {\small\texttt{RunBuildSystemTaskVerifier}} & Executes arbitrary build-system tasks and optionally validates command output against a regex with a configurable match-count threshold. \\
\addlinespace
Static analysis validators & {\small\texttt{NoFileErrorsVerifier}} & Invokes IDE static analysis on a file or directory and fails if any errors (or optionally warnings) are detected in files of the specified languages. \\
\bottomrule
\end{tabular}
\end{table}

\paragraph{LLM-judge metrics.}
Each trajectory is reviewed by LLM judges along five dimensions:
\begin{itemize}
    \item \textbf{\textsc{EndResult}} looks at the final outcome only, measuring completeness and fitness for purpose with respect to the user's request;
    \item \textbf{\textsc{InstructionCompliance}} checks whether the agent followed the user's explicit rules, requested steps, ordering constraints, and meta-instructions, independently of how good the final outcome was;
    \item \textbf{\textsc{Pitfalls}} catches the fixable behavioral failures that hinder the user: tool misuse, non-productive loops, premature completion, missing validation, workflow instability;
    \item \textbf{\textsc{Pleasantness}} captures user-facing interaction quality: clarity, accuracy, conciseness, helpfulness, and whether the session feels productive and non-disruptive;
    \item \textbf{\textsc{ToolCalls}} assesses tool-use quality: tool choice, argument correctness, success or failure, recovery from tool errors, and efficient versus futile calls.
\end{itemize}

This set is meant to be comprehensive for general coding agent evaluation (test generation uses a different metric set; see the repository).
Read together, the five dimensions separate the ways an agent can be good or bad: \textsc{EndResult} asks whether the assistant ultimately solved the task; \textsc{InstructionCompliance} asks whether it solved \emph{the requested} task under the user's constraints; \textsc{Pitfalls} exposes process-level instability; \textsc{ToolCalls} diagnoses failures in agent--environment interaction; and \textsc{Pleasantness} captures what it actually feels like to work with the assistant.

For every dimension the judge produces both a score and a written review (see the example prompt in Appendix~\ref{app:single-run-prompt}).
The review cites the trajectory evidence behind the score, so an aggregated report can preserve not only metric values but the reasons for them.
Scores live on a small ordinal scale (for \textsc{Pitfalls}, $0$ for miserable, $0.5$ for tolerable, $1$ for actually good), and each review ends with a handful of structured, reusable evidence lines of the form \texttt{Aspect | Severity | Evidence}.
When scores are aggregated across a run, the reviews are summarized into a single report; hyperlinks such as \texttt{[R1]} point back to the exact trajectory locations where a problem occurred, so a developer can jump straight to the evidence (see the example report in Appendix~\ref{app:single-run-report-gemini-31-pro}, and the condensed excerpt in Figure~\ref{fig:review-excerpt}).

\begin{figure}[t]
\begin{tcolorbox}[colback=gray!4,colframe=gray!50,title=\textbf{Excerpt: \textsc{EndResult} review, Gemini 3.1 Pro Preview},fonttitle=\small\bfseries,boxsep=2pt,left=4pt,right=4pt,top=2pt,bottom=2pt]
\small
\textbf{score mean:} \texttt{0.62}\\[2pt]
11 of 32 reviews judged the end result fit for purpose, typically where the change stayed narrowly scoped and was backed by passing targeted verification \dots\ The most common weakness was \emph{partial} completion rather than total failure: 7 reviews said required documentation or schema work was incomplete \dots\ 5 reviews found the delivered code state itself unusable or contradictory to the claimed outcome \dots\ and 7 more flagged unreliable validation/reporting, e.g.\ final reports claiming green tests despite recorded \texttt{spotless:check} and \texttt{enforcer:enforce} failures \texttt{[R3], [R10]}.
\end{tcolorbox}
\vspace{-4pt}
\caption{A condensed slice of one judge review. Each metric produces a short narrative grounded in numbered evidence pointers (\texttt{[R$k$]}) into the trajectory. The full report is in Appendix~\ref{app:single-run-report-gemini-31-pro}.}
\label{fig:review-excerpt}
\end{figure}

\paragraph{Deterministic telemetry.}
For every run we also collect hard statistics: interaction cost, cache hit rate, latency, termination reason, and generation-token throughput.
We compute tool-use telemetry too---the mean number of parallel tool calls, mean tool calls per chat, total tool-call count, and per-tool success rates.
These quantify efficiency and operational reliability independently of any subjective judgment.

Taken together, all of these metrics separate \emph{outcome} quality from \emph{process} quality, which matters because coding agents fail in different ways.
One may reach a good final answer inefficiently; another may follow instructions perfectly yet fail verification; a third may use tools incorrectly behind a plausible-looking response; a fourth may produce something useful through a thoroughly unpleasant interaction.
Scoring these dimensions separately gives diagnostic evidence rather than a single opaque number, which is exactly what you want when comparing general-purpose coding agents on realistic, multi-step work.

\subsection{Quality Index}
\label{sec:qi}

AgentLens reports the individual metric scores and a single aggregate quality index (QI).
The current index is the unweighted {mean} of the component scores, i.e., the five judge metrics together with formal verification:
\begin{equation}
    Q(a) = \frac{1}{|M|}\sum_{m \in M} s_m(a),
\end{equation}
where $a$ is an agent, $s_m(a) \in [0,100]$ is its aggregate score on metric $m$ (five listed above plus $\textsc{Formal}$).

We treat QI as a deliberately simple proxy for production quality.
The metrics already cover the main dimensions of user-visible coding-agent behavior---final usefulness, instruction following, tool use, recovery from failures, and interaction quality---so by Occam's razor the natural first index is just their unweighted average.
As a sanity check, we compared this index against the qualitative ``vibe check'' developers form from the reports, and the ranking is mostly intuitive: agents that feel reliable score high, while agents that need babysitting lose points for exactly the visible reasons one would expect---unfinished work, misleading validation, broken final states, or poor workflow control.

We also inspect correlations between metric aggregates across systems.
The raw correlations in Table~\ref{tab:absolute-metric-correlations} are uniformly positive, which is unsurprising: stronger systems tend to score higher on several dimensions at once.
Raw correlations therefore mostly confirm a broad ``quality factor'' rather than telling us which metrics are redundant.

\begin{table}[t]
\centering
\begin{minipage}[t]{0.48\textwidth}
\centering
\small
\setlength{\tabcolsep}{3.2pt}
\begin{tabular}{lrrrrrrr}
\toprule
 & QI & End & Form. & Instr. & Pitf. & Pleas. & Tools \\
\midrule
QI &  &  &  &  &  &  &  \\
End & \cellcolor{red!35}.91 &  &  &  &  &  &  \\
Form. & \cellcolor{red!27}.70 & \cellcolor{red!25}.65 &  &  &  &  &  \\
Instr. & \cellcolor{red!36}.95 & \cellcolor{red!34}.89 & \cellcolor{red!21}.56 &  &  &  &  \\
Pitf. & \cellcolor{red!38}.99 & \cellcolor{red!34}.90 & \cellcolor{red!26}.69 & \cellcolor{red!35}.93 &  &  &  \\
Pleas. & \cellcolor{red!37}.97 & \cellcolor{red!32}.85 & \cellcolor{red!25}.66 & \cellcolor{red!34}.89 & \cellcolor{red!36}.96 &  &  \\
Tools & \cellcolor{red!33}.88 & \cellcolor{red!26}.68 & \cellcolor{red!15}.40 & \cellcolor{red!32}.83 & \cellcolor{red!32}.85 & \cellcolor{red!33}.87 &  \\
\bottomrule
\end{tabular}
\caption{Pearson correlations between raw aggregate metrics ($n=15$). Abbreviations follow the metric names in Table~\ref{tab:leaderboard}.}
\label{tab:absolute-metric-correlations}
\end{minipage}\hfill
\begin{minipage}[t]{0.48\textwidth}
\centering
\small
\setlength{\tabcolsep}{3.2pt}
\begin{tabular}{lrrrrrr}
\toprule
 & End & Form. & Instr. & Pitf. & Pleas. & Tools \\
\midrule
End &  &  &  &  &  &  \\
Form. & \cellcolor{red!6}.16 &  &  &  &  &  \\
Instr. & \cellcolor{red!7}.19 & \cellcolor{blue!8}-.22 &  &  &  &  \\
Pitf. & \cellcolor{red!4}.10 & \cellcolor{blue!8}-.22 & \cellcolor{red!2}.06 &  &  &  \\
Pleas. & \cellcolor{blue!14}-.37 & \cellcolor{blue!28}-.74 & \cellcolor{blue!9}-.24 & \cellcolor{red!7}.17 &  &  \\
Tools & \cellcolor{blue!25}-.66 & \cellcolor{blue!16}-.42 & \cellcolor{blue!3}-.07 & \cellcolor{blue!10}-.27 & \cellcolor{red!6}.15 &  \\
\bottomrule
\end{tabular}
\caption{Pearson correlations between metrics normalized by quality index ($n=15$).}
\label{tab:relative-metric-correlations}
\end{minipage}
\end{table}

To factor out this common-quality effect, we also look at each metric divided by the quality index.
Because the index is a scaled sum of the components, the normalized metrics for a system sum to a constant, so Table~\ref{tab:relative-metric-correlations} describes metric \emph{profiles}: a positive correlation means two dimensions tend to take a large share of the total together (compensated elsewhere), while a negative correlation means they act as direct counterweights.

This profile view suggests the metrics carry largely non-redundant information.
Strong positive off-diagonal correlations are rare after normalization, so no two dimensions consistently occupy the same part of the quality profile.
The clearest counterweights are End Result versus Tool Calls ($r=-0.66$) and Formal Verification versus Pleasantness ($r=-0.74$): some systems earn a larger share of their score from final-state success or formal passes, while others earn it from tool-use discipline or user-facing interaction quality.
Instruction Compliance has only weak pairwise associations, which suggests it is not merely a proxy for final outcome or tool use.
These results support reporting the component metrics, not just the aggregate index.

\subsection{Side-by-Side Reviews}
\label{sec:sbs}

AgentLens also supports side-by-side comparison of two trajectories from the same scenario.
We reach for this when absolute scores are not enough to explain a practical difference between agents, and we use it for three things.
First, it exposes model-specific behavior---verbosity, literalness, recovery from errors, verification discipline.
Second, it helps compare strong and weak models beyond their leaderboard position.
Third, it detects regressions introduced by a change to the agent, the tool layer, or the runtime environment.

\paragraph{Pairwise LLM judge.}
For each task instance, the pairwise judge compares two agents that solved the same user request.
It receives both interaction trajectories together with the independent single-trajectory reviews for the dimension under comparison.
Those prior reviews are the primary evidence; the raw trajectories are consulted only to verify concrete details such as a specific tool call, an error message, or a final output.

\paragraph{Prompt and input structure.}
The prompt asks the judge to compare Agent~1 and Agent~2 with respect to a single performance dimension---instruction compliance, tool use, end-result quality, test quality, and so on.
It explicitly tells the judge to stay within that dimension, avoid introducing unrelated criteria, and avoid re-deriving the individual reviews from scratch.
This keeps the comparison anchored to the same evidence for both agents and reduces positional bias (Section~\ref{sec:pos_bias}).
Each pairwise prompt contains four conceptual blocks: (1) the trajectory of Agent~1; (2) the trajectory of Agent~2; (3) a prior single-agent review of Agent~1 for the target dimension; (4) the corresponding review of Agent~2.
Both trajectories correspond to the same task instance, and the dimension-specific prior reviews are the main basis for the decision (see the example prompt in Appendix~\ref{app:side-by-side-judge-prompt}).

\paragraph{Output format.}
The judge produces a short comparative explanation followed by structured evidence lines:
{\small
\begin{verbatim}
Aspect: <description> | Winner: A1/A2/Tie |
Severity: low/medium/high | Evidence: <short evidence>
\end{verbatim}
}
It then assigns an integer pairwise score. Note that severity here is \emph{relative}: it measures the gap between the two agents on an aspect, not how bad either is in absolute terms, so two agents that fail identically score a tie.

\subsection{Reliability and Bias Checks}

\subsubsection{Score Variance}
To estimate run-to-run variance, we repeated the same GLM-5.1 (self-hosted) configuration five times.
We picked GLM-5.1 on purpose as a mid-range model: strong enough to complete many scenarios, but shaky enough to actually expose variance in the benchmark.
Across the five runs, the quality index had mean $67.28$ and standard deviation $0.94$ on a $0$--$100$ scale.

Most of that variance came from formal verification rather than variance in judge scores.
A component-level decomposition attributed $60.5\%$ of index variance to formal verification, followed by Pitfalls at $18.1\%$ and End Result at $16.2\%$.
In other words, repeated runs mainly expose unstable task {completion}: whether the agent reaches a formally verified final state on borderline scenarios.

\begin{figure}[t]
\centering
\small
\setlength{\tabcolsep}{12pt}
\begin{tabular}{c|l|r}
\toprule
\shortstack{Formal verification\\passes} & Distribution & Count \\
\midrule
0/5 & \rule{0.5em}{0.7ex} & 1 \\
1/5 & \rule{1.5em}{0.7ex} & 3 \\
2/5 & \rule{2.5em}{0.7ex} & 5 \\
3/5 & \rule{2.0em}{0.7ex} & 4 \\
4/5 & \rule{2.0em}{0.7ex} & 4 \\
5/5 & \rule{7.5em}{0.7ex} & 15 \\
\bottomrule
\end{tabular}
\caption{Formal-verification stability across five repeated runs.}
\label{fig:formal-stability}
\end{figure}

Figure~\ref{fig:formal-stability} shows the same effect at the scenario level.
Of the 32 scenario--persona points, 15 passed formal verification in all five runs, one failed in all five, and the remaining 16 were flaky.
The benchmark's variance is thus concentrated in a small set of unstable points rather than distributed uniformly across all tasks, which makes repeated evaluation useful to
pinpoint the scenarios where an agent's behavior is nondeterministic or fragile.

\subsubsection{Side-by-Side Positional Bias}\label{sec:pos_bias}

To check whether side-by-side reviews are sensitive to the order in which the agents are presented, we ran an order-swap test.
For a pair of agents $A$ and $B$, we compare two prompts: $A$ as Agent~1 with $B$ as Agent~2 and vice versa.
Scores are oriented as the advantage of Agent~2 over Agent~1, so perfect antisymmetry would give $s(A,B) \approx -s(B,A)$.
For each metric we report the stable model difference, $\mathrm{effect}=(s(A,B)-s(B,A))/2$, and the order residual, $\mathrm{residual}=(s(A,B)+s(B,A))/2$.

\begin{table}[t]
\centering
\begin{minipage}[t]{0.63\textwidth}
\centering
\small
\setlength{\tabcolsep}{4pt}
\sisetup{round-mode=places, round-precision=2, table-format=-1.2}
\begin{tabular}{ll S S S S}
\toprule
Pair & Metric & {$s(A,B)$} & {$s(B,A)$} & {Effect} & {Residual} \\
\midrule
GLM-5.1 vs GLM-5.1 & End Result & 0.1938 & -0.2125 & 0.2032 & -0.0094 \\
GLM-5.1 vs GLM-5.1 & Instr. Compl. & -0.0187 & -0.0562 & 0.0188 & -0.0375 \\
GLM-5.1 vs GLM-5.1 & Pitfalls & 0.0000 & -0.0000 & 0.0000 & 0.0000 \\
GLM-5.1 vs GLM-5.1 & Pleasantness & 0.0063 & -0.0312 & 0.0188 & -0.0125 \\
GLM-5.1 vs GLM-5.1 & Tool Calls & -0.0437 & 0.0188 & -0.0313 & -0.0125 \\
\midrule
Opus 4.7 vs Haiku 4.5 & End Result & -0.3875 & 0.3500 & -0.3688 & -0.0188 \\
Opus 4.7 vs Haiku 4.5 & Instr. Compl. & -0.4188 & 0.3875 & -0.4032 & -0.0157 \\
Opus 4.7 vs Haiku 4.5 & Pitfalls & -0.4250 & 0.4313 & -0.4282 & 0.0032 \\
Opus 4.7 vs Haiku 4.5 & Pleasantness & -0.5000 & 0.5188 & -0.5094 & 0.0094 \\
Opus 4.7 vs Haiku 4.5 & Tool Calls & -0.4750 & 0.4125 & -0.4438 & -0.0313 \\
\bottomrule
\end{tabular}
\caption{Order-swap sanity check for side-by-side reviews. \emph{Effect} is the antisymmetric model-difference component; \emph{residual} is the remaining order-sensitive component.}
\label{tab:sbs-order-swap}
\end{minipage}\hfill
\begin{minipage}[t]{0.35\textwidth}
\centering
\small
\begin{tabular}{lrr}
\toprule
Metric & Self & Opp. \\
\midrule
Pleasantness & 11 & 0 \\
End Result & 7 & 3 \\
Tool Calls & 7 & 2 \\
Instruction Compliance & 3 & 2 \\
Pitfalls & 1 & 0 \\
\bottomrule
\end{tabular}
\caption{Self-preference flips across 157 task--metric comparisons for the GPT-5.5 and Sonnet~4.6 judges. ``Self'' means each judge preferred its own model family; ``Opp.'' means the reverse.}
\label{tab:self-preference}
\end{minipage}
\end{table}

Table~\ref{tab:sbs-order-swap} reports two such checks.
The GLM-5.1 pair is a close same-model comparison; the Opus--Haiku pair is a high-contrast one (Haiku was run in \emph{no-thinking} mode to widen the quality gap), and all five metrics preserve their direction after swapping sides.
In both settings the residuals are small in absolute value.
This does not prove that positional bias is absent, but it does suggest that order effects are secondary in these aggregate checks.

\subsubsection{Judge Self-Preference}

LLM judges can favor outputs from their own model family, especially when the systems being compared are close in quality.
To gauge how large this effect is in our setting, we compared the same GPT-5.5-vs-Sonnet-4.6 trajectories twice, once with GPT-5.5 as the judge and once with Sonnet~4.6.
These are two frontier models with close quality indices, so this is precisely the regime where self-preference should be the most visible.

The two judges picked different winners in $23\%$ of task--metric comparisons, and the disagreements were not symmetric: in $18\%$ of all comparisons each judge preferred the agent from its own family, while in $5\%$ each preferred the other agent. Half of this self-favoring bias comes from Pleasantness alone, which contributes 11 self-favoring flips and zero opposite-favoring ones; it makes sense that this happened with the most subjective metric (Table~\ref{tab:self-preference}). On the score scale, the average shift between the two judges is about $4\%$ of the full pairwise range.

In this comparison both judges still prefer the GPT-5.5 agent on average; the judge choice changes preference {strength} and borderline decisions, not the average winner; we report this as a caveat for close frontier-model comparisons. In the leaderboard experiments we keep GPT-5.4 fixed as the judge.


\subsection{Production Pipeline}
\label{sec:pipeline}

AgentLens is wired directly into our agent-development workflow.
A candidate version is run on the benchmark set, producing trajectories, formal-verification results, metric reviews, aggregate scores, and side-by-side comparisons against a baseline.

\paragraph{CI benchmark workflow.}
A continuous integration pipeline runs the automated benchmark evaluation.
It is triggered either on a schedule or manually by a researcher.
A preparation stage decides the evaluation mode (manual, nightly, or weekly).
The benchmark stage then runs the selected tasks in parallel across multiple language- and IDE-specific environments.
Each execution produces raw traces, logs, and agent-interaction dumps, stored as intermediate artifacts; after the run, these are aggregated and merged into a unified evaluation dataset.
From the merged dataset the pipeline computes the quantitative metrics and prepares a report containing both metric values and judge reviews (an example report is in Appendix~\ref{app:single-run-report-gemini-31-pro}).

\begin{figure}[t]
\centering
\includegraphics[width=0.6\textwidth]{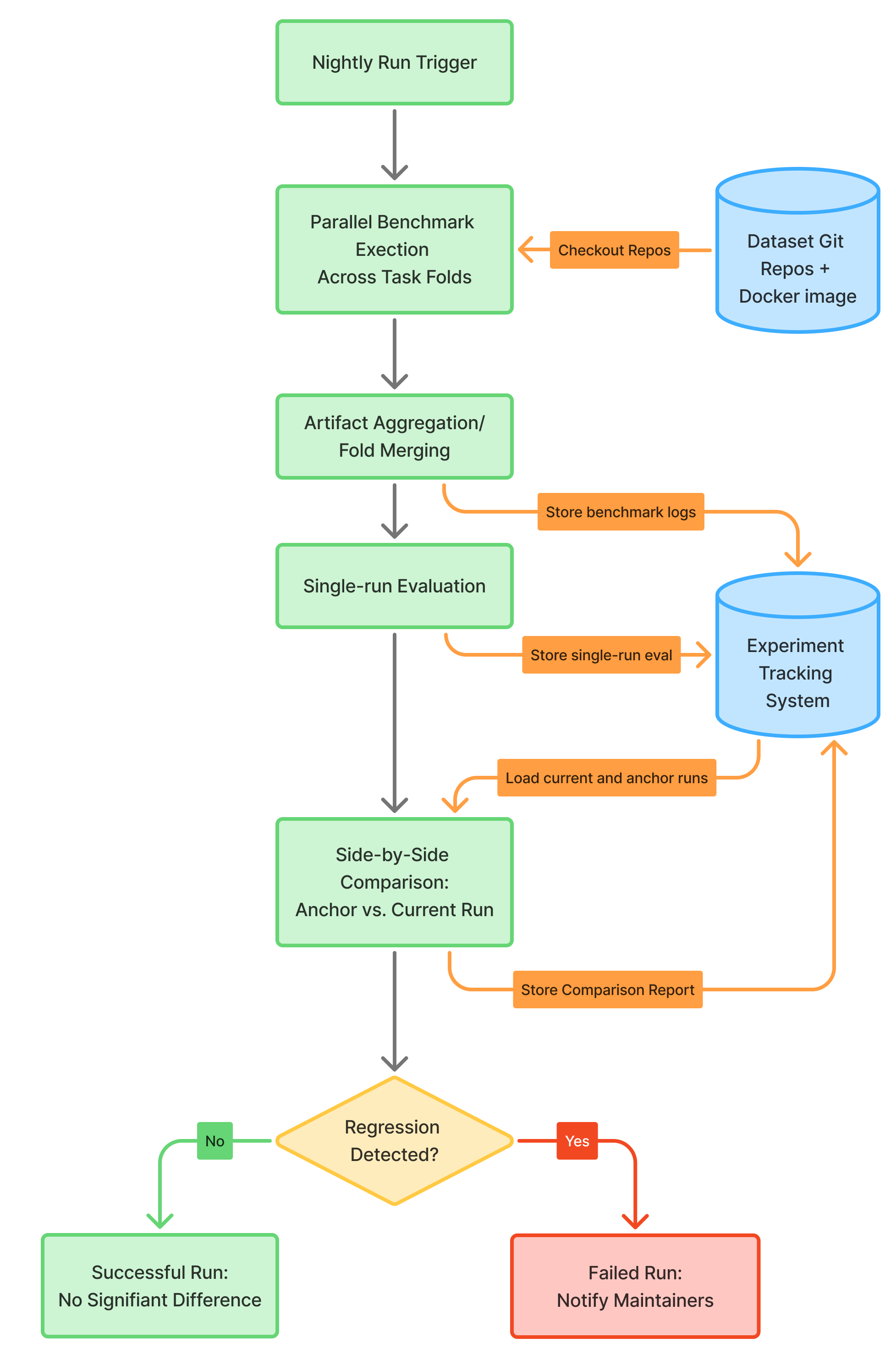}
\caption{Abstract CI workflow: benchmark execution, metric evaluation, experiment tracking, and regression detection via side-by-side comparison between the current and anchor runs.}
\label{fig:ci-benchmark-regression-workflow}
\end{figure}

Beyond absolute metrics, the pipeline detects regressions through side-by-side comparison.
The current run is compared against an \emph{anchor} run, which is supplied explicitly for manual experiments and chosen automatically for scheduled ones.
This comparison flags statistically significant performance changes between the current agent version and the reference (see the example report in Appendix~\ref{app:sbs-report-ds-flash-vs-pro}).
Detected degradations are treated as potential regressions and trigger a notification to maintainers.
Both reports are saved in the experiment-tracking system.
Figure~\ref{fig:ci-benchmark-regression-workflow} sketches the workflow.

Consider a real example: a race condition in parallel tool calling. It shows up as trajectory degradation, an abnormal termination reason together with worse ToolCalls and Pitfalls reviews. Since AgentLens exposes both the full interaction trajectory and the chat termination reason, the judges can attribute the regression to the underlying harness failure rather than merely noting a poor final answer. In this case the Pitfalls judge wrote, in the side-by-side review: \emph{``Tool and harness stability was mixed, but the most debug-relevant failures include Agent 2's explicit} \texttt{ConcurrentModificationException} \emph{crash in [C5]\dots''} (the full side-by-side review is boxed in Section~\ref{sec:qualitative}, Example~2).

\paragraph{Extensibility.}
The benchmark is not tied to a single agent or IDE.
Different CLI agents can be plugged in, and IDE-based assistants can be evaluated by implementing the required integration interfaces.
For IntelliJ IDEA plugins, the integration is done directly through a specialized \texttt{AgentEngine} interface; for CLI-based agents, the same contract is exposed through a lightweight Python adapter that translates benchmark requests into command-line invocations and returns the resulting trajectories.
Although this work focuses on Java, the benchmark adapts readily to other languages.
We open-source support for IntelliJ IDEA and PyCharm and plan to extend the integration to more IDEs and languages.

\section{Evaluation}
\label{sec:evaluation}

\subsection{Leaderboard}
\label{sec:leaderboard}

We evaluate AgentLens on Java coding-assistant tasks using two harnesses: our own \agentname\footnote{\url{https://explyt.ai/en}}(\abbrevagentname) and Claude Code\footnote{\url{https://claude.com/product/claude-code}}.
Table~\ref{tab:leaderboard} reports full-system performance, so a score reflects the model \emph{together with} its provider, agent loop, tool interface, and default execution policy---not the model in isolation.
All values are percentages, and rows are sorted by quality index.
All single-run reviews use GPT-5.4 as the judge.
We use high reasoning effort where supported, except where the table notes otherwise; OpenAI and Anthropic models use their official providers, GLM-5.1 (self-hosted) runs locally, and the remaining models are served through OpenRouter.

\begin{table}[!t]
\centering
\small
\setlength{\tabcolsep}{8pt}
\begin{tabular}{lllrrrrrr}
\toprule
Harness & Model & QI & Formal & End & Instr. & Pitfalls & Pleasant. & Tools \\
\midrule
\abbrevagentname & Opus 4.7 & 81.5 & 81.2 & 94.0 & 75.0 & 64.0 & 92.0 & 83.0 \\
Claude Code & Opus 4.7 (xhigh effort) & 76.2 & 81.2 & 91.0 & 69.0 & 58.0 & 78.0 & 80.0 \\
\abbrevagentname & GPT-5.5 & 73.0 & 75.0 & 78.0 & 70.0 & 55.0 & 80.0 & 80.0 \\
\abbrevagentname & Sonnet 4.6 & 70.2 & 81.2 & 66.0 & 62.0 & 56.0 & 73.0 & 83.0 \\
Claude Code & Sonnet 4.6 & 70.1 & 90.6 & 80.0 & 62.0 & 55.0 & 75.0 & 58.0 \\
\abbrevagentname & GLM-5.1 (self-hosted) & $67.3{\small\pm 0.9}$ & 72.5 & 71.8 & 57.4 & 53.2 & 75.8 & 73.0 \\
\abbrevagentname & DeepSeek V4 Flash & 64.8 & 87.5 & 73.0 & 50.0 & 50.0 & 66.0 & 62.0 \\
\abbrevagentname & DeepSeek V4 Pro & 64.1 & 87.5 & 66.0 & 53.0 & 48.0 & 69.0 & 61.0 \\
Claude Code & Haiku 4.5 & 63.6 & 84.4 & 78.0 & 58.0 & 48.0 & 58.0 & 55.0 \\
\abbrevagentname & GLM-5.1 & 60.6 & 80.6 & 60.0 & 53.0 & 48.0 & 61.0 & 61.0 \\
\abbrevagentname & Haiku 4.5 & 58.2 & 75.0 & 72.0 & 55.0 & 47.0 & 47.0 & 53.0 \\
\abbrevagentname & Gemini 3.1 Pro Preview & 54.5 & 71.9 & 62.0 & 52.0 & 41.0 & 50.0 & 50.0 \\
\abbrevagentname & Gemini 3 Flash & 52.7 & 75.0 & 66.0 & 48.0 & 41.0 & 39.0 & 47.0 \\
\abbrevagentname & MiniMax 2.7 & 50.8 & 70.0 & 52.0 & 47.0 & 38.0 & 43.0 & 55.0 \\
\abbrevagentname & Qwen 3.6 Plus & 49.1 & 65.6 & 58.0 & 45.0 & 39.0 & 42.0 & 45.0 \\
\abbrevagentname & Mimo V2.5 Pro & 45.1 & 62.5 & 50.0 & 38.0 & 32.0 & 36.0 & 52.0 \\
\abbrevagentname & Kimi K2.6$^{*}$ & 28.1 & 42.9 & 29.0 & 33.0 & 19.0 & 21.0 & 24.0 \\
\bottomrule
\end{tabular}
\caption{Leaderboard on the 32-trajectory Java benchmark. \textit{Formal} is the formal-verification pass rate; \textit{End}, \textit{Instr.}, \textit{Pitfalls}, \textit{Pleasant.}, and \textit{Tools} are LLM-judge metric means; \textit{QI} is the quality index. For selected agents we report the standard deviation of QI. $^{*}$Kimi~K2.6 had provider-side instability; results are included for completeness (see Section~\ref{sec:qualitative}).}
\label{tab:leaderboard}
\end{table}

\subsection{Model Behavior Case Study}\label{sec:model-behavior-study}
The leaderboard is the least interesting thing the benchmark produces.
The reviews are where the story is, so in this section we walk through the most revealing anomalies and close calls, using single-run and side-by-side reviews to understand \emph{why} models differ in reliability, tool use, instruction following, and raw API behavior.

\paragraph{Gemini 3.1 Pro Preview.}
Gemini 3.1 Pro Preview underperforms as an interactive coding agent relative to its frontier-model billing, and the report makes clear that the problem is agent-loop reliability rather than weak code generation.
It names workflow-control failures (``skipped or merged steps,'' ``premature edits/tests before approval''), validation failures (agents that ``claimed all checks passed despite actual failures''), and editing failures (``stale-patch or destructive-edit loops,'' ``brittle bulk replacements or sed that corrupted code'').
The point is not that Gemini cannot generate code---it is that high model capability did not translate into a dependable agent loop in this setup (full report in Appendix~\ref{app:single-run-report-gemini-31-pro}).

\paragraph{DeepSeek V4 Pro vs.\ DeepSeek V4 Flash.}
The near-tie between DeepSeek V4 Pro and Flash hides a more interesting trade-off than the QI suggests.
Pro is the more controlled agent: the side-by-side report shows a statistically significant instruction-compliance advantage, especially on step gating, required pauses, formatting, and file-boundary control (full report in Appendix~\ref{app:sbs-report-ds-flash-vs-pro}).
Flash trades compliance for progress---by relaxing step gating, formatting, and boundary constraints more often, it completes more work and more often lands in a usable, passing state within the fixed budget.
Higher throughput (more than $2\times$ in our run) reinforces the effect under a fixed task budget.
So the similar indices reflect genuinely different strategies: Pro earns quality through control and safer interaction, Flash through completion and follow-through.
A simple sum exposes the trade-off rather than hiding it.

\paragraph{\abbrevagentname Haiku vs.\ Claude Code Haiku.}
The gap between \agentname Haiku and Claude Code Haiku is likely explained in part by Claude Code Haiku's familiarity with the Claude Code harness, which may reflect prior training or evaluation exposure.
The difference shows up in the unusually low success rate of the \texttt{edit\_file} tool in \abbrevagentname relative to Claude Code ($0.66$ vs.\ $0.93$).
It is also visible in the pitfall analysis, where Claude Code is significantly favored (\texttt{metrics\_value}~$=0.1812$, $p=0.038$); the review puts the core distinction plainly, noting that \agentname ``more often makes poorly grounded or brittle changes and presents weaker verification.''
For the stronger Opus~4.7 model, though, \abbrevagentname already comes out ahead.

\paragraph{\abbrevagentname Opus vs.\ Claude Code Opus.}
This comparison splits along operational reliability versus explanatory and discovery strengths.
\agentname separates only narrowly on final outcomes---``no meaningful separation'' overall, with identical formal verification at $26$ vs.\ $26$---but the reviews still favor it on user-visible execution because it was ``more consistent at finishing changes cleanly and verifying them.''
Claude Code's high-contrast weakness is operational risk: judges cite repeated ``file/path and tool-contract errors,'' ``serious misses of explicit user constraints,'' a ``routing blast-radius regression,'' a ``leftover workspace mutation,'' and the only timeout.
Interestingly, the \texttt{edit\_file} success rate reverses the Haiku pattern here: Claude Code Opus is \emph{lower} than \abbrevagentname ($0.92$ vs.\ $1.00$).

\paragraph{GPT-5.5 vs.\ Opus 4.7.}
This pair shows a clean split between procedural compliance and semantic reliability.
GPT-5.5 often handles the explicit task protocol better---required message format, step boundaries, waiting for user confirmation, runner targeting, and compile/test closure.
But the reviews credit Opus with higher ``requirement coverage and final fitness,'' fewer ``likely route regressions'' or cases of moving a ``DB lookup ahead of the feature gating,'' and stronger validation discipline: GPT-5.5 was more often penalized for ``verification without covering changed branches,'' ``unsupported root-cause claims,'' ``missed affected tests,'' or marking a task done ``despite not reproducing the failure.''
In short, GPT-5.5 is better at following the task \emph{form}; Opus is better at preserving the task \emph{intent}.

\subsection{Qualitative Analysis}
\label{sec:qualitative}

The case study above used reviews to compare models. But the reviews are most clearly useful when they catch something a score alone would get \emph{wrong}. Here are two such cases; in both, the bare leaderboard number is, by itself, misleading, and only the written review recovers the truth.

\paragraph{A bottom-of-the-table score that is really a provider bug.}
Kimi K2.6 sits dead last in Table~\ref{tab:leaderboard}, which invites the obvious reading: it is a weak coding agent. But the ToolCalls review says otherwise: the dominant failure is not reasoning but a malformed-argument parsing problem on the OpenRouter side (the same JSON-wrapper mistake repeated across several tools), and once the arguments are flattened the agent does in fact use its tools successfully.

\begin{tcolorbox}[breakable,colback=gray!4,colframe=gray!55,title=\textbf{Example 1: Kimi K2.6 --- \textsc{ToolCalls} review (an OpenRouter tool-parser bug, not a capability gap)},fonttitle=\small\bfseries,boxsep=2pt,left=4pt,right=4pt,top=3pt,bottom=3pt]
\small
\textbf{ToolCalls\_Judge} \quad score mean: \texttt{0.24}

\smallskip
Malformed tool arguments were the dominant ToolCalls failure: 20 of 21 reviews reported at least one schema/argument parse error, and 17 of 21 described the same persistent wrapper mistake (\texttt{\{"": \{...\}\}} or deeper nesting) across tools such as \texttt{read\_file}, \texttt{search\_for\_text}, \texttt{run\_command}, \texttt{edit\_file}, \texttt{write\_file}, \texttt{list\_dir}, and GitHub helpers.

Recovery was often weak rather than immediate: 15 of 21 reviews explicitly say the agent retried near-identical malformed payloads after parser messages named the missing top-level fields, sometimes adding more nesting instead of flattening the JSON, which drove high futile churn and frequent timeouts.

Despite that, 20 of 21 reviews still recorded some successful tool use once flat JSON was used---e.g.\ repo reads/searches, GitHub issue or tag reads, edits/writes, or a passing test/configuration run---so the failure mode was usually contract confusion rather than tool unavailability.

The practical impact was substantial: 11 of 21 reviews explicitly report blocked end-to-end progress such as no successful edits, no build/test execution, or no PR/issue analysis completion, and 6 of 21 also note secondary inefficiency from overly broad scans or huge dumped outputs even when calls succeeded.
\end{tcolorbox}

\paragraph{A regression hiding behind a normal-looking answer.}
The second example comes from the nightly pipeline (Section~\ref{sec:pipeline}). During a weekly regression run, a race condition in parallel tool calling left little obvious trace in the final answer, but the side-by-side Pitfalls review pinned it to a concrete crash. Because AgentLens exposes the termination reason alongside the full trajectory, the judge could attribute the degradation to the harness---an explicit \texttt{ConcurrentModificationException}---rather than vaguely noting that the output had gotten worse.

\begin{tcolorbox}[breakable,colback=gray!4,colframe=gray!55,title=\textbf{Example 2: a parallel-tool-calling regression --- \textsc{Pitfalls} side-by-side review},fonttitle=\small\bfseries,boxsep=2pt,left=4pt,right=4pt,top=3pt,bottom=3pt]
\small
\textbf{Pitfalls\_Judge}\quad warning: \texttt{YES}\quad judge alert: \texttt{YES}\quad ($p=0.089$, metrics\_value $=0.152$)
\begin{PromptVerbatim}
The clearest pitfall pattern is verification accuracy: Agent 2 was better in 7 comparisons and worse in 2, with several serious cases where Agent 1 declared success after failed builds or incomplete checks ([C1], [C3], [C4], [C13]), while Agent 2’s misses here were lower-severity overclaims such as [C2] and [C11].

Tool and harness stability was mixed, but the most debug-relevant failures include Agent 2’s explicit ConcurrentModificationException crash in [C5] and Agent 1’s repeated malformed tool/edit calls and timeouts, including a 600s timeout in [C12] and broken edit invocations in [C11] and [C22]; across this group Agent 2 was better in 7 comparisons, worse in 4, with 3 ties.

On self-induced code breakage, Agent 2 was better in 9 comparisons and worse in 2: the strongest examples are Agent 1 breaking routing or Java file structure in [C3], [C7], [C19], and [C22], whereas Agent 2’s main losses were narrower regressions such as the bad setRequestor(...) edit in [C15] and a mock typing mistake in [C25].

On process discipline, Agent 2 was better in 9 comparisons and worse in 3, with repeated evidence that Agent 1 skipped required stop points or drifted from its own declared plan ([C4], [C6], [C15], [C21]), while Agent 2’s main opposing pattern was step-skipping of its own in [C16] and [C17].

By contrast, scope control and root-cause discipline leaned toward Agent 1: Agent 1 was better in 8 comparisons and worse in 5, driven by multiple high-severity cases where Agent 2 made speculative fixes without reproducing the bug or broadened contracts beyond the evidence ([C9], [C10]), plus requirement mismatches in [C18] and [C19].
\end{PromptVerbatim}
\end{tcolorbox}

For complete artifacts see Appendix~\ref{app:single-run-report-gemini-31-pro} and Appendix~\ref{app:sbs-report-ds-flash-vs-pro}.

\subsection{Correlation with Public Benchmarks}
\label{sec:aa-correlation}

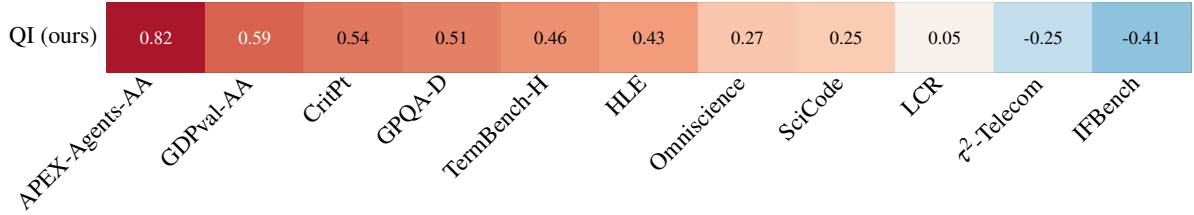
\begin{figure*}[t]
\centering
\pgfplotsset{
    colormap={RdBu}{
        rgb=(0.020,0.188,0.380)
        rgb=(0.129,0.400,0.674)
        rgb=(0.262,0.576,0.764)
        rgb=(0.572,0.772,0.870)
        rgb=(0.819,0.898,0.941)
        rgb=(0.968,0.968,0.968)
        rgb=(0.992,0.858,0.780)
        rgb=(0.956,0.647,0.509)
        rgb=(0.839,0.376,0.301)
        rgb=(0.698,0.094,0.168)
        rgb=(0.403,0.000,0.121)
    },
}
\begin{tikzpicture}
\begin{axis}[
    width=\textwidth,
    height=2.5cm,
    enlargelimits=false,
    colormap name=RdBu,
    point meta min=-1,
    point meta max=1,
    xtick={0,1,2,3,4,5,6,7,8,9,10},
    ytick={0.5},
    xticklabels={APEX-Agents-AA,GDPval-AA,CritPt,GPQA-D,TermBench-H,HLE,Omniscience,SciCode,LCR,$\tau^2$-Telecom,IFBench},
    yticklabels={QI (ours)},
    x tick label style={rotate=45, anchor=east, font=\small},
    y tick label style={font=\small},
    xmin=-0.5, xmax=10.5,
    ymin=-0.5, ymax=1.5,
]
\addplot[
    matrix plot*,
    mesh/cols=11,
    point meta=explicit,
] table [meta=z] {
x y z
0 1 0.821
1 1 0.591
2 1 0.536
3 1 0.509
4 1 0.460
5 1 0.427
6 1 0.273
7 1 0.255
8 1 0.050
9 1 -0.246
10 1 -0.409

0 0 0.821
1 0 0.591
2 0 0.536
3 0 0.509
4 0 0.460
5 0 0.427
6 0 0.273
7 0 0.255
8 0 0.050
9 0 -0.246
10 0 -0.409
};
    \node[font=\scriptsize,white] at (axis cs:0,0.5) {0.82};
    \node[font=\scriptsize,white] at (axis cs:1,0.5) {0.59};
    \node[font=\scriptsize,black] at (axis cs:2,0.5) {0.54};
    \node[font=\scriptsize,black] at (axis cs:3,0.5) {0.51};
    \node[font=\scriptsize,black] at (axis cs:4,0.5) {0.46};
    \node[font=\scriptsize,black] at (axis cs:5,0.5) {0.43};
    \node[font=\scriptsize,black] at (axis cs:6,0.5) {0.27};
    \node[font=\scriptsize,black] at (axis cs:7,0.5) {0.25};
    \node[font=\scriptsize,black] at (axis cs:8,0.5) {0.05};
    \node[font=\scriptsize,black] at (axis cs:9,0.5) {-0.25};
    \node[font=\scriptsize,black] at (axis cs:10,0.5) {-0.41};
\end{axis}
\end{tikzpicture}
\caption{Spearman rank correlation between our Quality Index (\textsc{QI}) and per-benchmark scores from Artificial Analysis, computed over the models selected from the IDE-agent leaderboard; the full inter-benchmark correlation matrix is in Figure~\ref{fig:benchmark-correlation}. Public benchmarks are sorted by correlation with our QI, highest to lowest.}
\label{fig:benchmark-correlation-one-row}
\end{figure*}

To place QI against the wider benchmark landscape, we computed Spearman rank correlations between our QI and every Artificial Analysis (AA)\footnote{\url{https://artificialanalysis.ai/}} evaluation that reports scores on the current generation of models.
Eleven evaluations qualified, including the agentic benchmarks APEX-Agents-AA, GDPval-AA, Terminal-Bench Hard, and $\tau^2$-Bench Telecom; ten cover all eleven IDE-agent models in Figure~\ref{fig:benchmark-correlation-one-row}, and APEX-Agents-AA covers seven.
AA also aggregates these into a single Intelligence Index (hereafter the \emph{AA composite})\footnote{\url{https://artificialanalysis.ai/methodology/intelligence-benchmarking}}, a weighted average designed to track overall model capability.
All ranks below are within these eleven models.

The QI column shows that
\emph{Mimo V2.5 Pro} and \emph{Gemini 3.1 Pro Preview} sit much higher on the AA composite than on our QI (rank gaps of $+4.9$ and $+3.8$; $90\%$ bootstrap CIs $[+3.4,+6.4]$ and $[+2.4,+5.1]$), consistent with optimization pressure toward AA-style academic and reasoning evaluations.
\emph{Claude Opus 4.7}, \emph{Claude Sonnet 4.6}, and \emph{DeepSeek V4 Flash} show the opposite, with rank gaps from $-3.4$ to $-3.9$ and CIs strictly negative: they handle long-horizon IDE work substantially better than the AA composite credits them for, with the gap concentrated on IFBench and $\tau^2$-Bench Telecom, where the Anthropic models in particular score well below their overall agentic quality.
\emph{GPT-5.5} and \emph{DeepSeek V4 Pro} sit at zero with CIs spanning the origin, the only two models for which the two rankings agree.

Two things follow.
First, QI tracks an axis of model quality that does not collapse to any single AA evaluation: the strongest individual correlate (APEX-Agents-AA, $\rho=0.82$) is itself an agentic benchmark, while reasoning- and instruction-following-oriented evaluations diverge sharply, including the \emph{negative} correlations of IFBench ($\rho=-0.41$) and $\tau^2$-Bench Telecom ($\rho=-0.25$).
Second, the rank-gap residuals look systematic rather than noisy: they split the cohort along a recognizable line between models trained for evaluation-style competence and models built for sustained tool use under realistic constraints.
The full inter-benchmark matrix in Figure~\ref{fig:benchmark-correlation} makes the same point from another angle: the public evaluations correlate strongly with one another, while our QI is the row that stands apart---it agrees with the agentic benchmarks and openly dissents from the reasoning- and instruction-following-oriented ones.

\newsavebox{\heatmapbox}
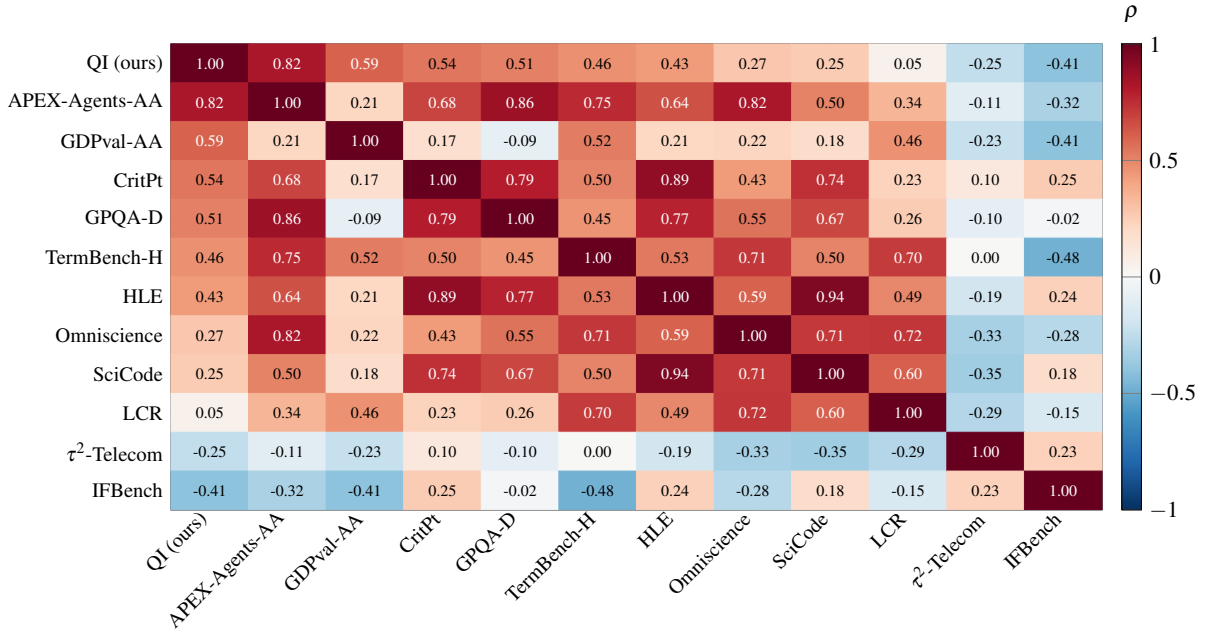
\begin{figure*}[t]
\centering
\pgfplotsset{
    colormap={RdBu}{
        rgb=(0.020,0.188,0.380)
        rgb=(0.129,0.400,0.674)
        rgb=(0.262,0.576,0.764)
        rgb=(0.572,0.772,0.870)
        rgb=(0.819,0.898,0.941)
        rgb=(0.968,0.968,0.968)
        rgb=(0.992,0.858,0.780)
        rgb=(0.956,0.647,0.509)
        rgb=(0.839,0.376,0.301)
        rgb=(0.698,0.094,0.168)
        rgb=(0.403,0.000,0.121)
    },
}
\begin{lrbox}{\heatmapbox}%
\begin{tikzpicture}
\begin{axis}[
    width=\textwidth,
    height=0.55\textwidth,
    enlargelimits=false,
    colormap name=RdBu,
    colorbar,
    point meta min=-1,
    point meta max=1,
    colorbar style={
        width=0.3cm,
        title={$\rho$},
    },
    xtick={0,1,2,3,4,5,6,7,8,9,10,11},
    ytick={0,1,2,3,4,5,6,7,8,9,10,11},
    xticklabels={QI (ours),APEX-Agents-AA,GDPval-AA,CritPt,GPQA-D,TermBench-H,HLE,Omniscience,SciCode,LCR,$\tau^2$-Telecom,IFBench},
    yticklabels={IFBench,$\tau^2$-Telecom,LCR,SciCode,Omniscience,HLE,TermBench-H,GPQA-D,CritPt,GDPval-AA,APEX-Agents-AA,QI (ours)},
    x tick label style={rotate=45, anchor=east, font=\small},
    y tick label style={font=\small},
    xmin=-0.5, xmax=11.5,
    ymin=-0.5, ymax=11.5,
]
\addplot[
    matrix plot*,
    mesh/cols=12,
    point meta=explicit,
] table [meta=z] {
x y z
0 11 1.000
1 11 0.821
2 11 0.591
3 11 0.536
4 11 0.509
5 11 0.460
6 11 0.427
7 11 0.273
8 11 0.255
9 11 0.050
10 11 -0.246
11 11 -0.409
0 10 0.821
1 10 1.000
2 10 0.214
3 10 0.679
4 10 0.857
5 10 0.750
6 10 0.643
7 10 0.821
8 10 0.500
9 10 0.342
10 10 -0.107
11 10 -0.321
0 9 0.591
1 9 0.214
2 9 1.000
3 9 0.173
4 9 -0.091
5 9 0.519
6 9 0.209
7 9 0.218
8 9 0.182
9 9 0.456
10 9 -0.232
11 9 -0.409
0 8 0.536
1 8 0.679
2 8 0.173
3 8 1.000
4 8 0.791
5 8 0.497
6 8 0.891
7 8 0.427
8 8 0.736
9 8 0.228
10 8 0.105
11 8 0.255
0 7 0.509
1 7 0.857
2 7 -0.091
3 7 0.791
4 7 1.000
5 7 0.451
6 7 0.773
7 7 0.545
8 7 0.673
9 7 0.255
10 7 -0.096
11 7 -0.018
0 6 0.460
1 6 0.750
2 6 0.519
3 6 0.497
4 6 0.451
5 6 1.000
6 6 0.528
7 6 0.711
8 6 0.497
9 6 0.703
10 6 0.002
11 6 -0.478
0 5 0.427
1 5 0.643
2 5 0.209
3 5 0.891
4 5 0.773
5 5 0.528
6 5 1.000
7 5 0.591
8 5 0.936
9 5 0.487
10 5 -0.187
11 5 0.236
0 4 0.273
1 4 0.821
2 4 0.218
3 4 0.427
4 4 0.545
5 4 0.711
6 4 0.591
7 4 1.000
8 4 0.709
9 4 0.715
10 4 -0.333
11 4 -0.282
0 3 0.255
1 3 0.500
2 3 0.182
3 3 0.736
4 3 0.673
5 3 0.497
6 3 0.936
7 3 0.709
8 3 1.000
9 3 0.601
10 3 -0.351
11 3 0.182
0 2 0.050
1 2 0.342
2 2 0.456
3 2 0.228
4 2 0.255
5 2 0.703
6 2 0.487
7 2 0.715
8 2 0.601
9 2 1.000
10 2 -0.292
11 2 -0.155
0 1 -0.246
1 1 -0.107
2 1 -0.232
3 1 0.105
4 1 -0.096
5 1 0.002
6 1 -0.187
7 1 -0.333
8 1 -0.351
9 1 -0.292
10 1 1.000
11 1 0.232
0 0 -0.409
1 0 -0.321
2 0 -0.409
3 0 0.255
4 0 -0.018
5 0 -0.478
6 0 0.236
7 0 -0.282
8 0 0.182
9 0 -0.155
10 0 0.232
11 0 1.000
};
    \node[font=\scriptsize,white] at (axis cs:0,11) {1.00};
    \node[font=\scriptsize,white] at (axis cs:1,11) {0.82};
    \node[font=\scriptsize,white] at (axis cs:2,11) {0.59};
    \node[font=\scriptsize,black] at (axis cs:3,11) {0.54};
    \node[font=\scriptsize,black] at (axis cs:4,11) {0.51};
    \node[font=\scriptsize,black] at (axis cs:5,11) {0.46};
    \node[font=\scriptsize,black] at (axis cs:6,11) {0.43};
    \node[font=\scriptsize,black] at (axis cs:7,11) {0.27};
    \node[font=\scriptsize,black] at (axis cs:8,11) {0.25};
    \node[font=\scriptsize,black] at (axis cs:9,11) {0.05};
    \node[font=\scriptsize,black] at (axis cs:10,11) {-0.25};
    \node[font=\scriptsize,black] at (axis cs:11,11) {-0.41};
    \node[font=\scriptsize,white] at (axis cs:0,10) {0.82};
    \node[font=\scriptsize,white] at (axis cs:1,10) {1.00};
    \node[font=\scriptsize,black] at (axis cs:2,10) {0.21};
    \node[font=\scriptsize,white] at (axis cs:3,10) {0.68};
    \node[font=\scriptsize,white] at (axis cs:4,10) {0.86};
    \node[font=\scriptsize,white] at (axis cs:5,10) {0.75};
    \node[font=\scriptsize,white] at (axis cs:6,10) {0.64};
    \node[font=\scriptsize,white] at (axis cs:7,10) {0.82};
    \node[font=\scriptsize,black] at (axis cs:8,10) {0.50};
    \node[font=\scriptsize,black] at (axis cs:9,10) {0.34};
    \node[font=\scriptsize,black] at (axis cs:10,10) {-0.11};
    \node[font=\scriptsize,black] at (axis cs:11,10) {-0.32};
    \node[font=\scriptsize,white] at (axis cs:0,9) {0.59};
    \node[font=\scriptsize,black] at (axis cs:1,9) {0.21};
    \node[font=\scriptsize,white] at (axis cs:2,9) {1.00};
    \node[font=\scriptsize,black] at (axis cs:3,9) {0.17};
    \node[font=\scriptsize,black] at (axis cs:4,9) {-0.09};
    \node[font=\scriptsize,black] at (axis cs:5,9) {0.52};
    \node[font=\scriptsize,black] at (axis cs:6,9) {0.21};
    \node[font=\scriptsize,black] at (axis cs:7,9) {0.22};
    \node[font=\scriptsize,black] at (axis cs:8,9) {0.18};
    \node[font=\scriptsize,black] at (axis cs:9,9) {0.46};
    \node[font=\scriptsize,black] at (axis cs:10,9) {-0.23};
    \node[font=\scriptsize,black] at (axis cs:11,9) {-0.41};
    \node[font=\scriptsize,black] at (axis cs:0,8) {0.54};
    \node[font=\scriptsize,white] at (axis cs:1,8) {0.68};
    \node[font=\scriptsize,black] at (axis cs:2,8) {0.17};
    \node[font=\scriptsize,white] at (axis cs:3,8) {1.00};
    \node[font=\scriptsize,white] at (axis cs:4,8) {0.79};
    \node[font=\scriptsize,black] at (axis cs:5,8) {0.50};
    \node[font=\scriptsize,white] at (axis cs:6,8) {0.89};
    \node[font=\scriptsize,black] at (axis cs:7,8) {0.43};
    \node[font=\scriptsize,white] at (axis cs:8,8) {0.74};
    \node[font=\scriptsize,black] at (axis cs:9,8) {0.23};
    \node[font=\scriptsize,black] at (axis cs:10,8) {0.10};
    \node[font=\scriptsize,black] at (axis cs:11,8) {0.25};
    \node[font=\scriptsize,black] at (axis cs:0,7) {0.51};
    \node[font=\scriptsize,white] at (axis cs:1,7) {0.86};
    \node[font=\scriptsize,black] at (axis cs:2,7) {-0.09};
    \node[font=\scriptsize,white] at (axis cs:3,7) {0.79};
    \node[font=\scriptsize,white] at (axis cs:4,7) {1.00};
    \node[font=\scriptsize,black] at (axis cs:5,7) {0.45};
    \node[font=\scriptsize,white] at (axis cs:6,7) {0.77};
    \node[font=\scriptsize,black] at (axis cs:7,7) {0.55};
    \node[font=\scriptsize,white] at (axis cs:8,7) {0.67};
    \node[font=\scriptsize,black] at (axis cs:9,7) {0.26};
    \node[font=\scriptsize,black] at (axis cs:10,7) {-0.10};
    \node[font=\scriptsize,black] at (axis cs:11,7) {-0.02};
    \node[font=\scriptsize,black] at (axis cs:0,6) {0.46};
    \node[font=\scriptsize,white] at (axis cs:1,6) {0.75};
    \node[font=\scriptsize,black] at (axis cs:2,6) {0.52};
    \node[font=\scriptsize,black] at (axis cs:3,6) {0.50};
    \node[font=\scriptsize,black] at (axis cs:4,6) {0.45};
    \node[font=\scriptsize,white] at (axis cs:5,6) {1.00};
    \node[font=\scriptsize,black] at (axis cs:6,6) {0.53};
    \node[font=\scriptsize,white] at (axis cs:7,6) {0.71};
    \node[font=\scriptsize,black] at (axis cs:8,6) {0.50};
    \node[font=\scriptsize,white] at (axis cs:9,6) {0.70};
    \node[font=\scriptsize,black] at (axis cs:10,6) {0.00};
    \node[font=\scriptsize,black] at (axis cs:11,6) {-0.48};
    \node[font=\scriptsize,black] at (axis cs:0,5) {0.43};
    \node[font=\scriptsize,white] at (axis cs:1,5) {0.64};
    \node[font=\scriptsize,black] at (axis cs:2,5) {0.21};
    \node[font=\scriptsize,white] at (axis cs:3,5) {0.89};
    \node[font=\scriptsize,white] at (axis cs:4,5) {0.77};
    \node[font=\scriptsize,black] at (axis cs:5,5) {0.53};
    \node[font=\scriptsize,white] at (axis cs:6,5) {1.00};
    \node[font=\scriptsize,white] at (axis cs:7,5) {0.59};
    \node[font=\scriptsize,white] at (axis cs:8,5) {0.94};
    \node[font=\scriptsize,black] at (axis cs:9,5) {0.49};
    \node[font=\scriptsize,black] at (axis cs:10,5) {-0.19};
    \node[font=\scriptsize,black] at (axis cs:11,5) {0.24};
    \node[font=\scriptsize,black] at (axis cs:0,4) {0.27};
    \node[font=\scriptsize,white] at (axis cs:1,4) {0.82};
    \node[font=\scriptsize,black] at (axis cs:2,4) {0.22};
    \node[font=\scriptsize,black] at (axis cs:3,4) {0.43};
    \node[font=\scriptsize,black] at (axis cs:4,4) {0.55};
    \node[font=\scriptsize,white] at (axis cs:5,4) {0.71};
    \node[font=\scriptsize,white] at (axis cs:6,4) {0.59};
    \node[font=\scriptsize,white] at (axis cs:7,4) {1.00};
    \node[font=\scriptsize,white] at (axis cs:8,4) {0.71};
    \node[font=\scriptsize,white] at (axis cs:9,4) {0.72};
    \node[font=\scriptsize,black] at (axis cs:10,4) {-0.33};
    \node[font=\scriptsize,black] at (axis cs:11,4) {-0.28};
    \node[font=\scriptsize,black] at (axis cs:0,3) {0.25};
    \node[font=\scriptsize,black] at (axis cs:1,3) {0.50};
    \node[font=\scriptsize,black] at (axis cs:2,3) {0.18};
    \node[font=\scriptsize,white] at (axis cs:3,3) {0.74};
    \node[font=\scriptsize,white] at (axis cs:4,3) {0.67};
    \node[font=\scriptsize,black] at (axis cs:5,3) {0.50};
    \node[font=\scriptsize,white] at (axis cs:6,3) {0.94};
    \node[font=\scriptsize,white] at (axis cs:7,3) {0.71};
    \node[font=\scriptsize,white] at (axis cs:8,3) {1.00};
    \node[font=\scriptsize,white] at (axis cs:9,3) {0.60};
    \node[font=\scriptsize,black] at (axis cs:10,3) {-0.35};
    \node[font=\scriptsize,black] at (axis cs:11,3) {0.18};
    \node[font=\scriptsize,black] at (axis cs:0,2) {0.05};
    \node[font=\scriptsize,black] at (axis cs:1,2) {0.34};
    \node[font=\scriptsize,black] at (axis cs:2,2) {0.46};
    \node[font=\scriptsize,black] at (axis cs:3,2) {0.23};
    \node[font=\scriptsize,black] at (axis cs:4,2) {0.26};
    \node[font=\scriptsize,white] at (axis cs:5,2) {0.70};
    \node[font=\scriptsize,black] at (axis cs:6,2) {0.49};
    \node[font=\scriptsize,white] at (axis cs:7,2) {0.72};
    \node[font=\scriptsize,white] at (axis cs:8,2) {0.60};
    \node[font=\scriptsize,white] at (axis cs:9,2) {1.00};
    \node[font=\scriptsize,black] at (axis cs:10,2) {-0.29};
    \node[font=\scriptsize,black] at (axis cs:11,2) {-0.15};
    \node[font=\scriptsize,black] at (axis cs:0,1) {-0.25};
    \node[font=\scriptsize,black] at (axis cs:1,1) {-0.11};
    \node[font=\scriptsize,black] at (axis cs:2,1) {-0.23};
    \node[font=\scriptsize,black] at (axis cs:3,1) {0.10};
    \node[font=\scriptsize,black] at (axis cs:4,1) {-0.10};
    \node[font=\scriptsize,black] at (axis cs:5,1) {0.00};
    \node[font=\scriptsize,black] at (axis cs:6,1) {-0.19};
    \node[font=\scriptsize,black] at (axis cs:7,1) {-0.33};
    \node[font=\scriptsize,black] at (axis cs:8,1) {-0.35};
    \node[font=\scriptsize,black] at (axis cs:9,1) {-0.29};
    \node[font=\scriptsize,white] at (axis cs:10,1) {1.00};
    \node[font=\scriptsize,black] at (axis cs:11,1) {0.23};
    \node[font=\scriptsize,black] at (axis cs:0,0) {-0.41};
    \node[font=\scriptsize,black] at (axis cs:1,0) {-0.32};
    \node[font=\scriptsize,black] at (axis cs:2,0) {-0.41};
    \node[font=\scriptsize,black] at (axis cs:3,0) {0.25};
    \node[font=\scriptsize,black] at (axis cs:4,0) {-0.02};
    \node[font=\scriptsize,black] at (axis cs:5,0) {-0.48};
    \node[font=\scriptsize,black] at (axis cs:6,0) {0.24};
    \node[font=\scriptsize,black] at (axis cs:7,0) {-0.28};
    \node[font=\scriptsize,black] at (axis cs:8,0) {0.18};
    \node[font=\scriptsize,black] at (axis cs:9,0) {-0.15};
    \node[font=\scriptsize,black] at (axis cs:10,0) {0.23};
    \node[font=\scriptsize,white] at (axis cs:11,0) {1.00};
\end{axis}
\end{tikzpicture}%
\end{lrbox}%
\resizebox{\linewidth}{!}{\usebox{\heatmapbox}}
\caption{Spearman rank correlation between our Quality Index
(\textsc{QI}) and per-benchmark scores from Artificial Analysis,
computed over the models selected from the IDE-agent leaderboard.
Kimi was excluded due to OpenRouter tool-parser failures during evaluation.
For GLM, we use the OpenRouter-served variant rather than a self-hosted deployment.
Public benchmarks are sorted by descending correlation with our QI.}
\label{fig:benchmark-correlation}
\end{figure*}

\section{Limitations}
\label{sec:limitations}

We note several limitations of our study. First, AgentLens evaluates a specific class of coding-agent tasks, not general model intelligence, and the present release is Java-only.

Second, Several evaluated models are accessed through third-party API providers (e.g.\ OpenRouter), so latency, routing, model versioning, and availability are outside our control.
The DeepSeek Pro-vs-Flash case in Section~\ref{sec:model-behavior-study} is a specific illustration for this effect: a $2\times$ throughput difference under a fixed task budget significantly affects how much work each agent can finish, which means some leaderboard differences reflect serving conditions as much as model quality. Third, benchmark cost is also non-trivial: a single run with Opus~4.7 can exceed \$100.

Finally, AgentLens was originally built around our own coding assistant, which may make some tasks or integration assumptions more natural for our system.
We mitigate this by releasing the benchmark and reporting full trajectory reviews, but broader external use is needed to assess neutrality.
The validity argument for our LLM judges (Section~\ref{sec:review-protocol}) currently rests on internal pairwise-annotation experience rather than a published agreement study; quantifying judge--human and judge--judge agreement on these long coding trajectories is an important direction for future work.

\section{Conclusion}
\label{sec:conclusion}

In this work, we have presented AgentLens, a benchmark that evaluates code agents through production-assessed trajectory reviews. It captures the user-visible quality dimensions that binary success metrics miss, and because every score comes with a written, evidence-linked review, it supports far more than ranking: diagnosis, feature evaluation, and nightly product-regression detection for a deployed coding assistant.

In future work, we plan to add integrations for more coding agents, including Codex and popular open-source agents such as OpenCode. Since AgentLens is pluggable, new CLI agents can be connected by implementing the required execution interface, and IDE-based assistants through the corresponding IDE integration layer. AgentLens already supports the same benchmark style across other JetBrains IDEs, including PyCharm, Rider, and WebStorm, and we plan to open-source these variants as the repository matures. The benchmark is also organized into task folds---test-writing tasks, for example, have their own fold with test-specific verifiers and judge metrics---which we likewise plan to release. This lets the benchmark grow by adding focused task families rather than by treating all coding-agent work as one homogeneous distribution.

Overall, we feel that the field of agentic coding benchmarks is undersaturated (albeit rapidly growing), and we hope that AgentLens fills a niche that is currently vacant.

\bibliographystyle{plainnat}
\bibliography{custom}

\appendix

\section{Example Single-Run Report}
\label{app:single-run-report-gemini-31-pro}

\begin{tcolorbox}[breakable,title=Gemini-3.1-pro-preview]

{\Large\bfseries Workflows\par}

{\large\bfseries TL;DR\par}

\begin{PromptVerbatim}
The agent performs best on narrowly scoped code changes: reviews repeatedly found successful refactors, targeted migrations, logging updates, and small interface changes when it stayed within scope and backed the work with focused tests. The main recurring weakness was partial or misleading delivery—documentation and schema work left incomplete, acceptance criteria covered only superficially, contract or behavior mismatches, and several tasks ending with a non-green or internally contradictory repository state despite claims that builds or static checks had passed. Operationally, about 70% of tasks reached formal verification, but validation was often narrower than requested, the edit-and-test workflow was unstable enough to trigger repeated patch/revert loops, and 4 of 32 conversations ended in timeout.
\end{PromptVerbatim}

{\large\bfseries number\_of\_points\par}

\begin{itemize}
\item chats\_count: \texttt{32}
\item agent\_scenarios\_count (tasks): \texttt{16}
\end{itemize}

{\large\bfseries formal\_verification\_result\par}

\begin{itemize}
\item value: \texttt{23}
\item success rate: \texttt{0.71875}
\end{itemize}

{\large\bfseries termination\_reason\par}

\begin{itemize}
\item User decided to finish the dialog: \texttt{28}
\item Waiting for the agent response timed out after 900s: \texttt{4}
\end{itemize}

{\large\bfseries Tool call success rates\par}

{\small
\begin{tabular}{lll}
\hline
Tool & successes/total & success rate \\
\hline
run\_configuration & \texttt{24/32} & \texttt{0.750} \\
edit\_file & \texttt{153/194} & \texttt{0.789} \\
pull\_request\_read & \texttt{8/8} & \texttt{1.000} \\
read\_file & \texttt{243/243} & \texttt{1.000} \\
... & ... & ... \\
\hline
\end{tabular}
}

{\large\bfseries EndResult\_Judge\par}

\begin{itemize}
\item score mean: \texttt{0.62}
\end{itemize}

\textbf{Review}

\begin{PromptVerbatim}
11 of 32 reviews judged the end result fit for purpose, typically where the change stayed narrowly scoped and was backed by passing targeted verification, such as the password refactors ([R1], [R27]), ApprovalService documentation pass ([R4]), Spring Boot 3.5.6 migration ([R9], [R20]), and the diagram, test-data, logging, bug-fix, and connector-endpoint updates in [R12], [R16], [R18], [R19], [R23], and [R24].

The most common weakness was partial completion rather than total failure: 7 reviews said required documentation or schema work was incomplete or insufficiently specific, especially missing Swagger model annotations or unresolved schema annotations in the TaskTemplate work ([R2], [R8], [R26], [R32]) and weaker-than-requested comment/diagram specificity in [R4], [R14], and [R29]; 4 more said tests or observability changes were too weak to enforce the ticket requirements, including missing sorting/edge-case assertions and non-structured logging that skipped exception paths ([R15], [R17], [R28], [R31]).

5 reviews found the delivered code state itself unusable or contradictory to the claimed outcome, due to unresolved compile/static-analysis errors in KafkaConnectService or test refactors ([R5], [R13], [R25], [R32]) or a claimed comparator fix that was not present in the final diff at all ([R10]).

7 reviews also flagged unreliable validation/reporting: some final reports claimed green tests or static analysis despite recorded failures like `spotless:check` and `enforcer:enforce` ([R3], [R10]), some ran only targeted tests instead of the requested build/module coverage ([R11], [R22]), and some left plugin-resolution warnings or unrelated failing tests unexplained or unproven as pre-existing ([R6], [R9], [R20]).

Public-contract and answer fidelity issues recurred too: one connector change implemented the wrong external route shape, `/topics/v2/connectors` instead of `/v2/connectors` ([R21]), and two Kafka-usage summaries overstated conclusions by ignoring `GROUP_ID_CONFIG`/`commitAsync()` semantics or omitting that producer usage was effectively test-only ([R7], [R30]).
\end{PromptVerbatim}

{\large\bfseries InstructionCompliance\_Judge\par}

\begin{itemize}
\item score mean: \texttt{0.52}
\end{itemize}

\textbf{Review}

\begin{PromptVerbatim}
The dominant InstructionCompliance failure was workflow control: at least 15 of 32 reviews reported skipped or merged steps, premature edits/tests before approval, or failure to stop at mandatory checkpoints, including jumping from Step 2 to Step 8, editing during reconnaissance, and combining post-Step results with the next step in the same message ([R5], [R6], [R7], [R12], [R15], [R16], [R18], [R19], [R23], [R29], [R30], [R31]).

A second recurring pattern was reply-format noncompliance in at least 14 of 32 reviews, especially missing `STARTING STEP N` headers, omitting the exact bridge phrase `Moving on to step 4`, starting messages with tool-call dumps, and failing required output contracts such as code blocks or the exact final phrasing ([R1], [R5], [R10], [R12], [R14], [R17], [R18], [R20], [R21], [R24], [R27], [R29], [R31]).

Required reconnaissance and reporting were also often incomplete: at least 9 of 32 reviews noted missing affected-file/test lists, merged per-comment plans, skipped method/branch coverage, missing environment/build-context details, or incomplete final summaries such as omitting changed-file counts and mandated completion lines ([R3], [R4], [R13], [R21], [R24], [R25], [R27], [R28]).

Verification compliance failed in at least 10 of 32 reviews: agents claimed all checks passed despite actual failures or missing tooling, skipped required call-tree analysis or debug-output steps, exceeded prescribed retry limits, or ran narrower checks than requested instead of the full build/test scope ([R3], [R10], [R11], [R13], [R16], [R17], [R19], [R20], [R22], [R23]).

Several reviews also show direct task-constraint breaches beyond workflow, including editing forbidden files/tests, changing serialization-related annotations, not using the mandated migration guide, and only partially implementing requested Swagger/model or typed-return changes ([R2], [R8], [R9], [R11], [R22], [R26], [R32]).
\end{PromptVerbatim}

{\large\bfseries Pitfalls\_Judge\par}

\begin{itemize}
\item score mean: \texttt{0.41}
\end{itemize}

\textbf{Review}

\begin{PromptVerbatim}
The most common pitfall was unverifiable or overstated validation, appearing in 13 of 32 reviews: agents reported clean checks after explicit failures or partial evidence, such as claiming success after `spotless`/`enforcer` failures ([R3]), saying a bug was fixed without a matching recorded diff ([R10]), or asserting compile/build success from ambiguous log tails or without running compile/build at all ([R18], [R22], [R28]).

Workflow discipline was the next major weakness in 11 of 32 reviews, with agents skipping required step gates or acting before approval—for example running `mvn clean verify` and editing before Step 3/4 confirmation ([R20]), implementing during reconnaissance before the plan was approved ([R19], [R31]), or jumping straight from early steps to `**STARTING STEP 6/8**` ([R7], [R15], [R30]).

Tooling misuse and state instability repeatedly amplified failures: 13 of 32 reviews showed invalid or noisy harness usage (`No configuration with name ...`, oversized dumped outputs, repeated `grep`/`cat`/`tail` log scraping) [[R2], [R3], [R14], [R16], [R23]], and 10 of 32 escalated into stale-patch or destructive-edit loops involving `Could not find old_text in file`, `sed -i`, `git reset --hard`, repeated `git restore`, broken workspaces, or timeouts [[R10], [R13], [R19], [R25], [R32]].

A separate 12 of 32 reviews involved incorrect problem framing—misreading code, requirements, or target scope—such as changing guarded control flow ([R6]), validating only unsorted role/permission sets despite an alphabetical requirement ([R17]), breaking auth by changing a known-good fixture ([R16]), or describing a `consumer.commitAsync()` path as non-committing ([R30]); 6 reviews also breached explicit constraints or dependencies by adding Swagger annotations in modules without `io.swagger.annotations`, touching forbidden `pom.xml`, editing forbidden tests, or making unsolicited commits [[R2], [R4], [R8], [R11], [R26], [R32]].
\end{PromptVerbatim}

{\large\bfseries Pleasantness\_Judge\par}

\begin{itemize}
\item score mean: \texttt{0.5}
\end{itemize}

\textbf{Review}

\begin{PromptVerbatim}
Pleasantness was most often damaged by trust issues: 16 of 32 reviews reported inaccurate, overstated, or poorly evidenced status claims, including saying builds/tests passed when tool output showed failures or no full build had run, and claiming files or behaviors were changed when they were not [[R3], [R5], [R10], [R11], [R18], [R22], [R28], [R30]].

A second recurring problem was noisy user-facing communication: 8 reviews exposed internal reasoning, meta-instructions, or raw tool chatter directly to the user, with examples such as visible `94>thought` blocks and self-talk like “Wait, let’s analyze...” that made the interaction feel unprofessional and confusing [[R6], [R7], [R16], [R23], [R25], [R28]].

Workflow discipline also regularly reduced smoothness: 14 reviews noted unnecessary confirmation loops, mismatched step labels, edits made before approval, or awkward step jumps that made progress feel unreliable even when the task eventually succeeded [[R1], [R3], [R9], [R18], [R19], [R29], [R31]].

Execution quality often compounded that friction: 12 reviews described thrashy or risky operations such as repeated failed edits, broad `sed` replacements, forbidden file changes, dependency probing, and timeouts/no final response, while 9 reviews separately reported incomplete delivery or scope drift that forced the user to do follow-up quality control [[R8], [R13], [R19], [R25], [R32]; [R4], [R11], [R12], [R18], [R21]].

When the interaction was pleasant, the common pattern was simple and consistent: concise step-based updates, repo-first investigation, limited targeted edits, transparent explanation of constraints, and concrete verification/results in the final report [[R1], [R14], [R17], [R21], [R24], [R26], [R27]].
\end{PromptVerbatim}

{\large\bfseries ToolCalls\_Judge\par}

\begin{itemize}
\item score mean: \texttt{0.5}
\end{itemize}

\textbf{Review}

\begin{PromptVerbatim}
Across the 32 reviews, the strongest recurring positive pattern was correct use of repo-aware discovery and scoped editing tools: many sessions quickly localized the target file or call site with `search_file_by_name`/`search_for_text`/MCP reads, inspected the relevant method with `read_file`, and landed a clean `edit_file` with no analysis failures (for example [R1], [R3], [R6], [R11], [R17], [R18], [R21], [R22], [R26], [R27]).

The main negative pattern was unstable editing in 11 reviews, typically caused by speculative changes, stale `old_text`, or unresolved dependencies/symbols rather than grounded file reads: repeated Swagger annotation failures in modules without the dependency ([R2], [R8], [R26], [R32]), wrong-file or wrong-API edits that had to be restored ([R5]), brittle bulk replacements or `sed` that corrupted code ([R13], [R19], [R25]), and edit loops where the claimed fix did not match the final file state ([R10]).

A second major weakness was inefficient tool selection and verification discipline: 15 reviews showed avoidable shell-heavy churn (`grep`/`cat`/`sed`, repeated `dependency:tree`, plugin/formatter invocations after success, or repeated test-file regeneration) instead of structured read/search/edit tools ([R2], [R5], [R7], [R10], [R12], [R15], [R16], [R20], [R21], [R23], [R29], [R30], [R31]), and 10 reviews had execution/validation mistakes such as guessed or wrong `run_configuration` targets, failure to read dumped outputs, single-test-only validation when a build was requested, or reporting completion without evidence that the verified state matched the claim ([R3], [R6], [R10], [R12], [R14], [R17], [R22], [R24], [R28], [R31]).

Harness-specific behavior was a consistent debugging signal: several agents correctly recovered from large-output temp-file dumps or flaky IDE-style runners by switching to `read_file` and direct Maven commands ([R1], [R4], [R6], [R7], [R17], [R21], [R23], [R28]), and some apparent IntelliJ/static-analysis failures were likely noise because Maven later succeeded ([R9], [R15], [R20], [R28]).

A smaller but important class of regressions came from unnecessary or disallowed workspace mutations, including editing forbidden files/tests or running risky VCS commands like `git reset --hard`, `git rebase`, `git clone`, and `git commit` that were unrelated to the user request ([R11], [R19], [R32], with lower-severity examples in [R4], [R10], [R21], [R23]).
\end{PromptVerbatim}

{\large\bfseries gen\_tokens\_to\_seconds\_ratio\par}

\begin{itemize}
\item value: \texttt{34.77}
\end{itemize}

{\large\bfseries tools\_call\_general\_stats\par}

\begin{itemize}
\item tools\_calls\_total: \texttt{1299}
\item tools\_calls\_per\_chat\_mean: \texttt{40.594}
\item tool\_calls\_in\_parallel\_mean: \texttt{1.237}
\end{itemize}

{\large\bfseries agent\_time\par}

\begin{itemize}
\item total: \texttt{14968}
\item response qtl (10/50/80/100): \texttt{10.00 / 52.50 / 152.00 / 900.00}
\item chat qtl (10/50/80/100): \texttt{142.10 / 309.50 / 875.20 / 1133.00}
\end{itemize}

{\large\bfseries agent\_price\par}

\begin{itemize}
\item total: ***
\item response qtl (10/50/80/100): \texttt{*/*/*/*}
\item chat qtl (10/50/80/100): \texttt{*/*/*/*}
\end{itemize}

{\large\bfseries Metadata\par}

\begin{itemize}
\item run name: \texttt{gemini-31-pro-preview-*}
\item scenario tag: \texttt{workflows}
\item language: \texttt{java}
\item model\_info: \texttt{Google/gemini-3.1-pro-preview}
\item plugin\_hash: \texttt{***}
\item judge: \texttt{gpt-5.4}
\item evaluation price: \texttt{***}
\item time of creation: \texttt{***}
\item dataset rel path: \texttt{***.json}
\item dataset\_config\_hash: \texttt{***}
\item path to IDEA dumps: \texttt{***}
\item launch args: \texttt{***}
\end{itemize}

\end{tcolorbox}

\section{Example Side-by-Side Report}
\label{app:sbs-report-ds-flash-vs-pro}

\begin{tcolorbox}[breakable,title=DeepSeek-v4-pro vs. DeepSeek-v4-flash]

\section*{Workflows}

\subsection*{TL;DR}

\begin{PromptVerbatim}
The main operational split is that Agent 1 was more trustworthy about protocol, diagnostic grounding, and not overstating validation, while Agent 2 more often got code and tests to a buildable, fully executed state but also had several serious cases of claiming success on weak or misleading verification evidence. On the outcome itself, the reviews are genuinely mixed rather than decisive: Agent 2 had more wins on completeness, fallback work, and executed verification, but Agent 1 had several of the more consequential wins where Agent 2 missed explicit requested behavior, introduced contract or fixture errors, or overfit the task; the aggregate outcome edge toward Agent 2 is small and not statistically reliable (p=0.117). Instruction-following shows a similar pattern, with Agent 2 somewhat stronger on sequencing, message format, and actually finishing required test steps, while Agent 1 was better on scope control in several important cases and more consistently followed special procedural requirements; the formal verification count was tied at 28-28, which is worth noting but only as a rough heuristic rather than strong proof of correctness. Operationally, Agent 2 was much cheaper and faster ($9.13 vs $21.20, 10,973s vs 18,987s, both statistically significant), and its much higher command/run success rates align with the reviews showing steadier completion, whereas Agent 1’s extra cost and slower pace came with more timeouts/exceptions and poorer run-configuration success, even though its final reporting was often the more dependable of the two.
\end{PromptVerbatim}

\subsubsection*{Total number of points (chats): 32}

\subsection*{formal\_verification\_result}

\begin{itemize}
    \item alert: no
    \item warning: no
    \item p-value: \texttt{1.0}
    \item metrics:
    \begin{itemize}
        \item agent 1: \texttt{28}
        \item agent 2: \texttt{28}
    \end{itemize}
\end{itemize}

\subsection*{termination\_reason}

\begin{itemize}
    \item alert: no
    \item warning: YES
    \item common: User decided to finish the dialog (26), Waiting for the agent response timed out after 900s (1)
    \item only\_run1: Agent iteration failed with an unexpected exception: Agent loop failed with an unexpected error: Cannot reach provider OpenAI-compatible (1), Scenario timed out after 1200s (3), Waiting for the agent response timed out after 900s (1)
    \item only\_run2: User decided to finish the dialog (5)
\end{itemize}

\subsection*{EndResult\_Judge}

\begin{itemize}
    \item alert: no
    \item warning: YES
\end{itemize}

\textbf{Review}:

\begin{PromptVerbatim}
Across requirement fit and scope/completeness, the results are mixed: A2 leads in 20 Ck while A1 leads in 11, with the most important A2 wins coming from delivered or more complete fallback/test work in [C32], [C29], [C22], and [C26], but A1 has notable high-severity wins where A2 missed explicit requested behavior or overfit the task, especially [C25], [C21], and [C14].

On correctness and behavior preservation, neither side dominates overall (A2 wins 10 Ck, A1 wins 9), but the most consequential failures are split: A1 has high-severity wins in [C25] and [C21] because A2 introduced a wrong `/v2` route contract and left an invalid username fixture, while A2 has high-severity wins in [C11], [C29], and [C4] because A1 changed control-flow semantics, left a broken buildable state, or omitted required `501` handling.

Verification/build evidence favors A2 in 9 Ck versus 2 for A1, with the strongest gaps in [C29], [C32], and [C22] where A2 delivered passing builds or suites while A1 was broken, unrun, or incomplete; A1’s main counterexample is [C19], where A2’s final verification depended on extra skip flags rather than the agreed full command.

For documentation/report accuracy, the pattern is again mixed (A2 wins 8 Ck, A1 wins 7), with A1’s most significant advantages in [C9], [C14], [C21], and [C24] where A2 had factual errors, overclaimed a fix, or used overly absolute wording, while A2’s strongest wins in [C10], [C12], [C16], and [C22] come from catching inconsistent prerequisites, invalid diagrams, inaccurate completion claims, or unsupported rationale.

Safety/logging hygiene shows only a small signal: A1 has the main substantive win in [C1] by removing a risky full-request log, while [C2] is a tie on safe logging choices.

\end{PromptVerbatim}

\begin{itemize}
    \item judge alert: YES
\end{itemize}

\textbf{Scoring}
\begin{itemize}
    \item p-value: \texttt{0.117}
    \item metrics\_value (from [-1..1]): \texttt{0.1375}
\end{itemize}

\subsection*{InstructionCompliance\_Judge}

\begin{itemize}
    \item alert: no
    \item warning: YES
\end{itemize}

\textbf{Review}:

\begin{PromptVerbatim}
Across the step-gating and sequencing evidence, Agent 2 has the stronger record: A2 wins 14 comparisons in this group, while A1 wins 10 and 9 are ties; the clearest cases are [C1], [C2], [C29], [C30], and [C32] for A2, versus [C13], [C20], and [C21] for A1.

On message-format compliance, Agent 2 also has the edge, winning 9 items versus 8 for Agent 1, with many of A2’s wins being more severe, including high-severity header/STOP-structure cases in [C26] and [C27]; Agent 1’s main counterexamples are [C20], [C21], [C23], and [C28].

Agent 1 is stronger on scope control and special-procedure requirements: in scope/allowed-files behavior A1 wins 7 items versus 6 for A2, with major A1 wins in [C5], [C8], and [C15] but major A2 wins in [C7] and [C18]; in tools/diagnostics/language/protocol specifics A1 wins 8 items versus 3 for A2, driven by call-tree and mandated-phrase cases such as [C14] and [C19].

Test-execution/verification evidence favors Agent 2 overall, with 6 wins for A2 versus 2 for A1 and 2 ties, especially in [C21], [C22], [C29], and [C30] where A2 is credited for tighter test scope or actually completing the required verification/reporting stage.

\end{PromptVerbatim}

\begin{itemize}
    \item judge alert: YES
\end{itemize}

\textbf{Scoring}
\begin{itemize}
    \item p-value: \texttt{0.855}
    \item metrics\_value (from [-1..1]): \texttt{-0.0187}
\end{itemize}

\subsection*{Pitfalls\_Judge}

\begin{itemize}
    \item alert: no
    \item warning: YES
\end{itemize}

\textbf{Review}:

\begin{PromptVerbatim}
Across workflow/protocol and verification/build-state reliability, Agent 1 had the stronger record: in workflow-related proofs A1 was better in 14 entries and worse in 4, with major separations in [C10] and [C24], and in verification/reporting A1 was better in 10 entries and worse in 9, including high-severity cases where A2 treated failing or stale runs as success in [C4], [C14], [C19], and [C20].
Agent 1 also led on logic/contract correctness, with 7 wins versus 5 losses in that group; the most consequential A2 misses were fixes that did not address the reported path in [C14], a filtering/pagination regression in [C18], and fallback handling errors in [C26], while A2’s main counterexamples were A1’s routing miss in [C3] and semantics break in [C11].
Agent 2’s clearest advantages were narrower and more operational: it was better on edit stability in 8 entries versus 7 losses, especially where A1 destabilized files in [C3], [C25], and [C29], and it was also better on some scope-control items such as [C1], [C7], [C17], [C21], and [C26], though A2 had severe scope violations of its own in [C8] and [C18].
Tool/harness hygiene was mixed rather than decisive, but several A1 losses there were tied to repeated command/schema mistakes or noisy reruns (for example [C13], [C25], [C29], and [C31]), while A2 had its own notable tool-side regressions in [C20] and [C32].
Overall, the extracted proofs point to Agent 1 being more reliable on protocol, diagnostic grounding, and behavior correctness, while Agent 2 was somewhat steadier on avoiding file corruption and some scope-follow-through but with less favorable high-severity verification failures.
\end{PromptVerbatim}

\begin{itemize}
    \item judge alert: YES
\end{itemize}

\textbf{Scoring}
\begin{itemize}
    \item p-value: \texttt{0.109}
    \item metrics\_value (from [-1..1]): \texttt{-0.125}
\end{itemize}

\subsection*{Pleasantness\_Judge}

\begin{itemize}
    \item alert: no
    \item warning: YES
\end{itemize}

\textbf{Review}:

\begin{PromptVerbatim}
Across the strongest evidence, Agent 1 leads on interaction hygiene: it wins verification/reporting trust in 11 Ck and loses it in 5, including high-severity cases where Agent 2 gave contradictory or misleading validation narratives in [C4], [C12], and [C14]; Agent 1 also leads on structure/protocol in 6 Ck versus 3 losses, with the biggest gaps in [C21], [C20], and [C23] where Agent 2 skipped or broke the required step format.

Agent 2’s clearest advantage is responsiveness and end-to-end completion: it wins responsiveness/alignment in 8 Ck and loses 1, with major cases in [C26] and [C32], and it wins completion/final usefulness in 4 Ck versus 4 losses, including strong delivery wins in [C29] and [C32].

On operational smoothness, the split is close but important: Agent 1 wins 11 Ck and loses 11, yet Agent 2 owns several high-severity wins in [C25] and [C29], while Agent 1’s biggest smoothness wins come from avoiding disruptive behavior in [C12], [C13], and [C16].

Boundary-respecting behavior is mixed rather than one-sided: Agent 1 wins scope-discipline in 3 Ck and loses 3, with both agents drawing criticism for touching forbidden files or constraints (for example ([C5]/[C6]/[C8] versus [C7]/[C28])).

\end{PromptVerbatim}

\begin{itemize}
    \item judge alert: YES
\end{itemize}

\textbf{Scoring}
\begin{itemize}
    \item p-value: \texttt{0.317}
    \item metrics\_value (from [-1..1]): \texttt{-0.0875}
\end{itemize}

\subsection*{ToolCalls\_Judge}

\begin{itemize}
    \item alert: no
    \item warning: YES
\end{itemize}

\textbf{Review}:

\begin{PromptVerbatim}
Across the proof groups, the largest repeated differences are in edit stability and verification reliability: Agent 2 is favored in edit/repository stability in 7 comparisons and loses 8, but Agent 1’s worse cases include several high-severity corruptions or malformed edits such as [C18], [C29], and [C5], while Agent 2’s high-severity losses are concentrated in [C8] and [C16].

Verification and runner use lean toward Agent 2 in 11 comparisons versus 6 for Agent 1, with Agent 2’s strongest wins coming from correctly scoped or completed test/build execution in [C21], [C25], [C26], and [C32]; Agent 1’s strongest wins in this area are mostly about Agent 2 masking Maven failures with shell pipelines in [C19] and [C20].

Exploration efficiency is mixed but slightly favors Agent 1 overall, with 13 wins for Agent 1 versus 10 for Agent 2; Agent 1’s wins are mostly about avoiding broad or empty searches (for example [C1], [C2], [C6], [C14]), while Agent 2’s wins come from more targeted narrowing in search-heavy tasks such as [C23] and [C24] and less repeated configuration probing in [C21]–[C22].

Scope/context/completion shows the sharpest directional signal toward Agent 2: Agent 2 wins 5 comparisons and loses 2, including high-severity cases where Agent 1 edited forbidden files or failed to complete the final requested tool work ([C7], [C26], [C32]), while Agent 1’s counterexamples are narrower scope violations by Agent 2 in [C5] and [C8].

Miscellaneous argument/path/output handling is broadly noisy on both sides, with many ties and small wins split both ways; the most notable non-tie cases are Agent 1 being better when Agent 2 makes unknown-tool or bad-command mistakes ([C20], [C30]), and Agent 2 being better when Agent 1 makes invalid read ranges or unhelpful file-type reads ([C9], [C22]).

\end{PromptVerbatim}

\begin{itemize}
    \item judge alert: YES
\end{itemize}

\textbf{Scoring}
\begin{itemize}
    \item p-value: \texttt{0.92}
    \item metrics\_value (from [-1..1]): \texttt{-0.0125}
\end{itemize}

\subsection*{setups}

\begin{itemize}
    \item alert: no
    \item warning: no
    \item difference: \texttt{unchecked}
\end{itemize}

\subsection*{Inference metrics}

\begin{center}
\resizebox{\linewidth}{!}{
\begin{tabular}{l l l l l l}
\hline
Metric & p-value & run 1 & run 2 & alert & warning \\
\hline
price & \texttt{0.0011} & \texttt{21.1996815} & \texttt{9.129628199999997} & \textcolor[HTML]{CC2222}{\textbf{YES}} & YES \\
time & \texttt{0.0291} & \texttt{18987} & \texttt{10973} & \textcolor[HTML]{CC2222}{\textbf{YES}} & YES \\
gen tokens to seconds ratio & \texttt{0.0} & \texttt{24.45} & \texttt{53.66} & \textcolor[HTML]{CC2222}{\textbf{YES}} & YES \\
tools calls total & \texttt{0.3646} & \texttt{1302} & \texttt{1548} & no & no \\
generation tokens & \texttt{0.1379} & \texttt{464235} & \texttt{588772} & no & no \\
cache hit mean ratio & \texttt{0.9946} & \texttt{0.6296} & \texttt{0.6245} & no & no \\
tool calls in parallel mean & -- & \texttt{1.644} & \texttt{1.621} & no & no \\
\hline
\end{tabular}
}
\end{center}

\subsection*{Tool call success rates}

\begin{center}
\resizebox{\linewidth}{!}{
\begin{tabular}{l l l l l l l}
\hline
Tool & run 1 success & run 2 success & run 1 total & run 2 total & alert & warning \\
\hline
pull\_request\_review\_write & \texttt{1.0} & \texttt{0} & \texttt{1} & \texttt{0} & no & no \\
read\_terminal & \texttt{1.0} & \texttt{0.8333333333333334} & \texttt{1} & \texttt{6} & no & no \\
edit\_file & \texttt{0.943089430894309} & \texttt{0.9084967320261438} & \texttt{123} & \texttt{153} & no & no \\
get\_configurations & \texttt{1.0} & \texttt{0.9672131147540983} & \texttt{64} & \texttt{61} & no & no \\
search\_file\_by\_name & \texttt{1.0} & \texttt{0.9733333333333334} & \texttt{126} & \texttt{150} & no & no \\
file\_structure & \texttt{1.0} & \texttt{0.9761904761904762} & \texttt{15} & \texttt{42} & no & YES \\
find\_class\_source & \texttt{0.9393939393939394} & \texttt{0.9230769230769231} & \texttt{33} & \texttt{13} & no & YES \\
list\_dir & \texttt{1.0} & \texttt{1.0} & \texttt{33} & \texttt{46} & no & no \\
write\_file & \texttt{1.0} & \texttt{1.0} & \texttt{30} & \texttt{22} & no & no \\
pull\_request\_read & \texttt{1.0} & \texttt{1.0} & \texttt{13} & \texttt{7} & no & no \\
issue\_read & \texttt{1.0} & \texttt{1.0} & \texttt{9} & \texttt{9} & no & no \\
web\_fetch & \texttt{1.0} & \texttt{1.0} & \texttt{4} & \texttt{10} & no & no \\
get\_release\_by\_tag & \texttt{1.0} & \texttt{1.0} & \texttt{3} & \texttt{1} & no & no \\
developer\_platform\_search & \texttt{1.0} & \texttt{1.0} & \texttt{1} & \texttt{2} & no & no \\
similar\_search & \texttt{1.0} & \texttt{1.0} & \texttt{1} & \texttt{2} & no & no \\
search\_repositories & \texttt{1.0} & \texttt{1.0} & \texttt{1} & \texttt{1} & no & no \\
search\_by\_name & \texttt{0} & \texttt{0.0} & \texttt{0} & \texttt{1} & no & no \\
search\_for\_text & \texttt{0.994413407821229} & \texttt{1.0} & \texttt{179} & \texttt{202} & no & no \\
read\_file & \texttt{0.9871323529411765} & \texttt{0.9948979591836735} & \texttt{544} & \texttt{588} & no & no \\
run\_command & \texttt{0.9411764705882353} & \texttt{0.9933774834437086} & \texttt{51} & \texttt{151} & no & YES \\
run\_configuration & \texttt{0.5757575757575758} & \texttt{0.9178082191780822} & \texttt{66} & \texttt{73} & \textcolor[HTML]{CC2222}{\textbf{YES}} & YES \\
find\_usages & \texttt{0.0} & \texttt{0.3333333333333333} & \texttt{4} & \texttt{6} & \textcolor[HTML]{CC2222}{\textbf{YES}} & YES \\
get\_me & \texttt{0} & \texttt{1.0} & \texttt{0} & \texttt{1} & no & no \\
list\_releases & \texttt{0} & \texttt{1.0} & \texttt{0} & \texttt{1} & no & no \\
\hline
\end{tabular}
}
\end{center}

\subsection*{Metadata}

\textbf{run1}
\begin{itemize}
    \item model\_info: \texttt{deepseek-v4-pro}
    \item plugin\_hash: \texttt{***}
    \item run\_name: \texttt{DeepSeek\_v4\_pro-*}
\end{itemize}

\textbf{run2}
\begin{itemize}
    \item model\_info: \texttt{deepseek-v4-flash}
    \item plugin\_hash: \texttt{***}
    \item run\_name: \texttt{DeepSeek\_v4\_flash-*}
\end{itemize}

\textbf{judge}
\begin{itemize}
    \item api\_calls: \texttt{***}
    \item cached\_tokens: \texttt{***}
    \item completion\_tokens: \texttt{***}
    \item model: \texttt{OpenAI/gpt-5.4}
    \item prompt\_tokens: \texttt{***}
    \item total\_price\_usd: \texttt{***}
\end{itemize}

\end{tcolorbox}

\section{Example Judge Prompts}

\subsection{Pitfalls single-run judge prompt} \label{app:single-run-prompt}
\begin{tcolorbox}[breakable]
\small

\textbf{System prompt.}
You are a coding expert and an objective critic. Your work will be checked and judged later on. Vagueness leads to your death.

\vspace{0.5em}
\textbf{User prompt.}

\begin{PromptVerbatim}
<INTERACTION_HISTORY>

Above you have a history of interactions between a user simulator and an agent.
This history contains additional system info and system messages that we have
passed to the simulator during the conversation. Some parts were truncated to
shorten your prompt and the user simulator's prompt. For example, you might see
"remaining characters are omitted only from the User Simulator's prompt..."
message. It means that an original transcript was truncated only after the task
was completed.

Tool call responses might contain lists of errors that occurred during the tool
call: if such a list is empty, it means that there were no errors (e.g., empty
compilation_errors section for edit_file tool). The last user message might not
be present in the history: either because the user decided to terminate the chat
or because a time limit was reached; the reason is unknown.

Note that this interaction was collected in a benchmark environment, that differs
from real life situations: for example, when the agent catches a timeout, it means
that it has probably hit the benchmark time limit.

Your task is to judge the agent, not the simulator. So if the task is not achieved
due to user simulator failures, it should not affect the score you give. Treat the
simulator's actions as a given, even if they seem suboptimal, and judge the agent
based on how well it responds within that context.

There might be an <agent_actions_summary> section in their chat. We append it to
help the simulator understand what the agent did. As the simulator-agent
conversation progresses, this info gets updated and appended to the last agent
message sent to the simulator. All previous information gets dropped from the
previous messages though, so you don't see them in the given history. This system
info consists of:
 - a section with a list of compilation errors from IDE analysis (project errors);
   if empty, there are no compilation errors in the project;
 - all the code diffs made by the agent (maybe empty);
 - the most recent cmd calls (maybe empty).

Note that we use a simulator instead of a real user: it behaves based on character
traits given to it.

Below we've gathered a final summary of the agent's actions and state of the project:
<agent_actions_summary>
<AGENT_TRACE>
</agent_actions_summary>

The chat has terminated for the following reason:
<termination_reason>
<TERMINATION_REASON>
</termination_reason>

<Instructions>
Your task is to act as a precise and analytical judge of an agent-user interaction.
Judge the agent (not the simulator) on the performance dimension 'Pitfalls' only.

Identify and diagnose fixable pitfalls in the agent's behavior that hindered the user.
This is about instability patterns and self-sabotage dynamics (often
harness/tooling/workflow), not end-result quality.

Extract pitfalls using this rubric. For each distinct pitfall, provide:
- Category: Logic/Code faults | Tool/Harness faults | Process/Interaction faults | Something else
- Severity: low/medium/high
- Frequency: once/repeated
- Evidence: a short quote (tool name + error text, or a brief message fragment)
- Mechanism: 1 short clause on why this happened (root cause)

Counting rule (mandatory):
- One behavioral cluster (e.g., a loop of identical tool calls) counts as at most
  one pitfall; if it repeats with the same root cause, mark it as Frequency=repeated.

Pitfall burden total (mandatory):
- severity points: low=1, medium=3, high=7
- frequency multiplier: once=x1, repeated=x2
- Total = sum(severity_points * frequency_multiplier) across distinct pitfalls

Write pitfalls as structured lines so they can be reused later:
`Pitfall: <free text> | Category: ... | Severity: ... | Frequency: ... |
Evidence: ... | Mechanism: ...`

Also include one line: `BurdenTotal=<Total>`.
</Instructions>

<Scoring>
Score should be a number from [0, 0.5, 1], where:
- 0 means miserable performance,
- 0.5 means tolerable,
- 1 means actually good.
</Scoring>

Response consists of 2 parts:
- <Review>, 2 to 5 sentences narrative focused on the metric. After the narrative,
  add 1 to 6 short structured evidence lines when applicable:
  `Aspect: <free text> | Severity: low/medium/high | Evidence: <short quote/tool/error>`
  Make Aspect-lines concrete and reusable: they are the main guide for later aggregation.
  Severity is absolute (not delta-based).
- <Score>, the sole final number. Must be one number and nothing else.

Template:
<Review>
[2-5 sentences]
Aspect: ... | Severity: ... | Evidence: ...
...

<Score>
[0|0.5|1]

Provide your response in the requested format:
\end{PromptVerbatim}
\end{tcolorbox}

\subsection{Pitfalls side-by-side judge prompt} \label{app:side-by-side-judge-prompt}
\begin{tcolorbox}[breakable]
\small

\textbf{System prompt.}
You are a coding expert and an objective critic. Your work will be checked and judged later on. Vagueness leads to your death.

\vspace{0.5em}
\textbf{User prompt.}

\begin{PromptVerbatim}
<agent1_trajectory>
<AGENT_1_INTERACTION_HISTORY>
</agent1_trajectory>

<agent2_trajectory>
<AGENT_2_INTERACTION_HISTORY>
</agent2_trajectory>

Above are two chat transcripts (each between a user simulator and an agent) for the same task, one per agent, delimited by <agent1_trajectory> and <agent2_trajectory> tags. Some parts were truncated to shorten your prompt and the user simulator's prompt. For example, you might see "remaining characters are omitted **only from the User Simulator's prompt**..." message. It means that an original transcript was truncated only after the task was completed.

<agent1_review>
<PRECOMPUTED_SINGLE_TRAJECTORY_REVIEW_FOR_AGENT_1>
</agent1_review>

<agent2_review>
<PRECOMPUTED_SINGLE_TRAJECTORY_REVIEW_FOR_AGENT_2>
</agent2_review>

The two blocks above, delimited by the <agent1_review> and <agent2_review> tags, are precomputed single-trajectory reviews from another pipeline. They are for the same task instance, one per agent, both focused on the same performance dimension. They were computed independently from the raw transcripts.
The performance dimension name for these reviews is 'Pitfalls'.

<Instructions>
Your task is to compare Agent 1 and Agent 2 on the SAME task instance with respect to the performance dimension 'Pitfalls'.
Use the precomputed reviews (<agent1_review>, <agent2_review>) as the primary input to your comparison to avoid positional bias.
Do NOT introduce new aspects beyond what is present in those reviews, and do NOT revise the individual assessments inside them.
You may consult the raw trajectories ONLY to verify concrete points mentioned in the reviews (e.g., a specific tool call, error message, or key claim), or get the general context.
Stay strictly on the performance dimension 'Pitfalls'; ignore unrelated aspects.
Do not propose fixes; only characterize differences with a focus on issues.

Inside the <Comparison> section, end your text with a short structured breakdown (1-6 lines) so a summarizer can weigh severity:
`Aspect: <free text> | Winner: A1/A2/Tie | Severity (Significance * Degree of Divergence): low/medium/high | Evidence: <specific quote/tool/error>`
Start with aspects where Agent 2 is a winner.
Keep aspects specific to 'Pitfalls', and keep evidence concrete and extra short. You should assess the severity of the observed differences accurately. Severity should be determined as the product of the aspect's Significance and the Degree of Divergence between agents on that aspect. If both agents exhibit the same failure mode, severity should be low even if that failure is critical in absolute terms. Severity is about the delta between agents, not how bad the situation is in absolute terms. For instance, the significance of the aspect 'breaking the repo to a non-recoverable state' is critical. If both agents break the repo, but one breaks one module and another breaks two modules, the divergence is little. So the overall Severity of this difference is low.

The Metric-Specific Instructions are described below. When identifying aspects to compare, use <agent1_review> and <agent2_review> as your primary source. Other aspects are assessed by other judges, don't steal their work.
<Metric-Specific Instructions>
Focus on identifying and diagnosing fixable failures in the agents' behavior that hinder the user (mostly harness/tooling/workflow issues). This is about instability patterns and self-sabotage dynamics, not outcome quality.

POSITION-INVARIANCE (MANDATORY):
- The *order* of presentation must not affect your judgement.
- However, you must keep identity consistent: any pitfall you list must be clearly attributed to Agent 1 or Agent 2 and must not "flip" mid-comparison.
- Write your analysis so that if the trajectories were swapped, the reasoning would stay the same and only the final score sign would flip.

EVALUATION PROTOCOL (follow in this order):
1) Extract pitfalls separately for Agent 2 and Agent 1 using the same rubric. For each pitfall provide: Category, Severity (low/medium/high), Frequency (once/repeated), and a short evidence quote (tool name + error text, or a brief message fragment).
2) Compute pitfall burden totals Total_1 and Total_2 (formula below).
3) Only after step (2), map (Total_1 vs Total_2) to a score using the deterministic rules below.

CATEGORIES (use consistent taxonomy):
- Logic/Code faults: demonstrable wrong code/logic or missing validation that risks wrong output.
- Tool/Harness faults: tool misuse, wrong args, inability to recover from tool errors, patching mistakes, broken edit_file usage, etc.
- Process/Interaction faults: misunderstanding requirements, non-productive loops, poor communication, premature "done", ignoring constraints.
- One behavioral cluster (e.g., a loop of identical tool calls) must be counted as at most one distinct pitfall. If the same tool or process error appears two or more times with the same root cause, it must be labeled as "repeated" and counted as a single pitfall with a frequency multiplier, not as multiple distinct pitfalls.

KEEP EXAMPLES CONCRETE:
(tool misuse, loops, missing validation, partial completion, placeholders, poor communication, self-contradiction, premature completion, failure to verify).
Emphasize causal mechanisms (why the failure happens), not just outcomes.

SCORING (SYMMETRIC, FORMULA-BASED):
Compute a pitfall burden score for each side and derive the pairwise score from the delta.

A) PITFALL BURDEN SCORING PER AGENT
For each distinct pitfall you listed in step (1), assign points:
- low = 1
- medium = 3
- high = 7
Frequency multiplier:
- once = x1
- repeated/looping/compounding = x2
So each pitfall contributes (severity_points * frequency_multiplier).
Sum to get totals: Total_1 (Agent 1), Total_2 (Agent 2).

B) DELTA (deterministic)
Delta = Total_2 - Total_1.
(Positive Delta means Agent 1 is better; negative Delta means Agent 2 is better.)

C) MAP DELTA TO FINAL SCORE (integer in [-5..5])
Use |Delta| only to choose the magnitude; the sign is determined solely by which agent is worse.
Magnitude table:
- |Delta| <= 3  -> 0
- 4..6          -> 1
- 7..10         -> 2
- 11..15        -> 3
- 16..22        -> 4
- >=23          -> 5
Sign rule (STRICT):
- If Delta > 0 (Agent 1 better): score MUST be negative.
- If Delta < 0 (Agent 2 better): score MUST be positive.
- If Delta == 0: score = 0.

OUTPUT REQUIREMENT (to enforce symmetry):
At the end of your narrative (before the structured Aspect lines), include a single line:
`Totals: A1=<Total_1>, A2=<Total_2>, Delta=A2-A1=<Delta> => Score=<score>`

CRITICAL:
- The final score MUST be computed ONLY from this Totals/Delta line using the mapping rules above.
- Never mix up agent identities: A1 refers to Agent 1 trajectory, A2 refers to Agent 2 trajectory.
- Do not use any other "overall impression". If your narrative suggests something else, adjust the pitfall list and totals until they match.
These numbers must be consistent with your pitfall list and with the mapping rules above.
</Metric-Specific Instructions>

Scoring guidelines:
Reserve the |4-5| score range for stark contrasts: one agent remains stable and correct, while the other exhibits cascading errors that destabilize the repository, gets stuck, or leaves the result completely unverified. When differences are clear but non-critical, use lower scores. Below is an example specific for Pitfalls and ToolCalls dimension, where the difference is clear, but not critical:
    - Agent N made a clear tool-handling error: after list_dir, which dumped the output to a temporary file, it attempted to read a different path via read_file, receiving the warning "Virtual file does not exist," which created unnecessary friction. Agent M handled the same "dumped to file" pattern correctly, read the proper path, and proceeded without errors.
    Aspects:
    Temp-file tool handling correctness | Winner: AM | Severity: low | Evidence: AN read_file -> "Virtual file does not exist ..." after list_dir
    Non-productive detour after tool error | Winner: AM | Severity: low | Evidence: AN made an additional search_file_by_name call after the failure
    Score given: 1 or -1 depending on who is N and M
Below is an example specific for Pitfalls and ToolCalls dimension, where both agents had issues, but "manual rollback" seemed more critical:
    - Agent N introduced a workflow-stopping repository break: multiple edit_file operations corrupted test code (e.g., turning an assignment into a stray queryConfig( line), causing cascaded compile/IDE errors ("Cannot resolve symbol...", "';' expected") and requiring manual recovery via git checkout -- .... Agent M made some incorrect assumptions and minor requirement drift, but did not corrupt files or require rollback, and kept changes localized.
    Aspects:
    Manual rollback needed to recover | Winner: AM | Severity: high | Evidence: AN git checkout -- ...ConfigControllerTest.java / ...ConfigFileControllerTest.java
    Cascading compile breakage from tool misuse | Winner: AM | Severity: medium | Evidence: AN widespread "Cannot resolve method 'queryConfig'" after edits
    File corruption via edits (repo inconsistent) | Winner: AM | Severity: medium | Evidence: AN ConfigControllerTest.java shows broken queryConfig( line; IDE errors "Cannot resolve symbol 'result'", "';' expected"
    Misleading intermediate claim (corrected). | Winner: AN | Severity: medium | Evidence: AM premature call-site claim later corrected via .queryConfig( search
    Score given: 3 or -3 depending on who is N and M

Response consists of two parts:
- <Comparison>, 2 to 5 sentences PLUS the breakdown lines at the end. Avoid primacy and framing bias. Do not start the comparison with "Agent 1 ..." or "Agent 2 ...". First describe the concrete difference in behavior in neutral, role-free terms (e.g., "One trajectory shows repeated edit failures due to missing required arguments, while the other applies edits without tool errors").
Only after stating the difference, explicitly assign Winner: A1/A2/Tie in the structured breakdown. Keep the narrative symmetric and avoid framing one agent as the default reference point.
- <PairwiseScore>, an integer in [-5, 5]. Use negative when Agent 1 is better, positive when Agent 2 is better.
  -5 means Agent 1 is MUCH better and the difference would feel very severe for the user. 0 means tie/indistinguishable. +5 means Agent 2 is MUCH better and the difference would feel very severe for the user. So the score is symmetric this way. When setting the score, mentally check the sign: negative for Agent 1 win, positive for Agent 2 win.

Template:
<Comparison>
[2-5 sentences comparison] +
Aspect: ... | Winner: ... | Severity: ... | Evidence: ...
...

<PairwiseScore>
[-5..5]
</Instructions>

Provide your response in the requested format:
\end{PromptVerbatim}

\end{tcolorbox}

\section{Benchmark Tasks}

\paragraph{Task categories.}\label{app:task-clusters}

Table~\ref{tab:covered-cluster-examples} shows examples of anonymized production-usage clusters covered by benchmark scenarios. The descriptions are generated from anonymized task summaries and are included to illustrate the type of real coding work represented in the benchmark.

\begin{table*}[t]
\centering
\small
\setlength{\tabcolsep}{4pt}
\begin{tabular}{p{0.18\linewidth}p{0.24\linewidth}p{0.50\linewidth}}
\toprule
Task family & Representative tags & Cluster description \\
\midrule
Unit testing and test refactoring & java, testing, junit, springboot, mockito, unittesting, maven, refactoring & The user is working on a Java project, typically using Spring Boot, and needs to establish or improve unit testing. They are focused on writing and refactoring tests with frameworks like JUnit and Mockito, often mocking dependencies such as repositories or services. Common activities include generating comprehensive test suites, optimizing existing tests, and integrating tests into build systems like Maven or Gradle. The work emphasizes ensuring proper test coverage, handling edge cases, and maintaining clean, maintainable test code. \\
\addlinespace
Legacy database-logic migration & database, migration, springboot, java, liquibase, sql, jpa, debugging & The user is migrating legacy database logic, typically SQL or Oracle functions, into Java code within a Spring Boot application. This involves designing and implementing JPA entities, repositories, and service layers to replace the database-side logic. A key component is using Liquibase for database schema changes and migrations, often to support the new Java structures or seed test data. Common challenges include debugging migration errors, resolving constraint violations, and ensuring correct dependency injection in the Spring context. \\
\addlinespace
API documentation and DTO cleanup & documentation, java, refactoring, springboot, openapi, dto, codecomments, swagger & The user is working on a Java Spring Boot application, often dealing with legacy code or existing codebases, and aims to improve the code quality and clarity for future maintenance. A primary focus is on enhancing API documentation---particularly through OpenAPI/Swagger annotations---and ensuring consistent JSON field naming, such as enforcing camelCase in DTOs. This involves systematic tasks such as adding or refining JavaDoc comments, standardizing exception handling, and removing outdated annotations. The overarching goal is to prepare the code for refactoring or to align with modern API standards while maintaining a clear, documented codebase. \\
\bottomrule
\end{tabular}
\caption{Examples of production-usage clusters represented by benchmark scenarios.}
\label{tab:covered-cluster-examples}
\end{table*}

\end{document}